\documentclass[10pt,journal,cspaper,compsoc]{IEEEtran}

\usepackage{float}
\usepackage{graphicx}
\usepackage{amsmath}
\usepackage{amsthm}
\usepackage{amssymb}
\usepackage{booktabs}
\usepackage[lined]{algorithm2e}
\usepackage[pagebackref=true,breaklinks=true,letterpaper=true,colorlinks,bookmarks=false]{hyperref}
\usepackage{placeins}

\newcommand{\R}[1]{\mathbb{R}^{#1}}
\newcommand{\RR}[2]{\mathbb{R}^{#1 \times #2}}

\newcommand{\pr}{\mbox{Pr}}

\DeclareMathOperator*{\argmax}{arg\!\max}
\DeclareMathOperator*{\argmin}{arg\!\min}

\newcommand{\refEq}[1]{(\ref{#1})}
\newcommand{\refFig}[1]{Figure~\ref{#1}}

\newcommand{\refSec}[1]{Section~\ref{#1}}

\newcommand{\refAlg}[1]{Algorithm~\ref{#1}}
\newcommand{\refTab}[1]{Table~\ref{#1}}

\def\bfc{{\boldsymbol{c}}}

\def\bfu{{\boldsymbol{u}}}
\def\bfv{{\boldsymbol{v}}}
\def\bfw{{\boldsymbol{w}}}
\def\bfx{{\boldsymbol{x}}}
\def\bfy{{\boldsymbol{y}}}

\def\bfB{{\boldsymbol{B}}}
\def\bfC{{\boldsymbol{C}}}

\def\bfI{{\boldsymbol{I}}}

\def\bfK{{\boldsymbol{K}}}

\def\bfM{{\boldsymbol{M}}}

\def\bfR{{\boldsymbol{R}}}
\def\bfS{{\boldsymbol{S}}}
\def\bfT{{\boldsymbol{T}}}

\def\bfW{{\boldsymbol{W}}}

\def\bfZ{{\boldsymbol{Z}}}

\def\bfone{{\boldsymbol{1}}}

\begin{document}

\title{MonoCap: Monocular Human Motion Capture using a CNN Coupled with a Geometric Prior}

\author{Xiaowei~Zhou,~\IEEEmembership{}
        Menglong Zhu, ~\IEEEmembership{}
        Georgios Pavlakos, ~\IEEEmembership{}
        Spyridon Leonardos,~\IEEEmembership{}
        \\Konstantinos G. Derpanis~\IEEEmembership{}
        and~Kostas~Daniilidis,~\IEEEmembership{Fellow,~IEEE}
\IEEEcompsocitemizethanks{\IEEEcompsocthanksitem X.Z. is with the College of Computer Science, Zhejiang University, China. M.Z., G.P., S.L. and K.D. are with Computer and Information Science Department and GRASP Laboratory, University of Pennsylvania, USA. K.G.D. is with the Department of Computer Science, Ryerson University, Canada. \protect\\
E-mail: xzhou@cad.zju.edu.cn}
\thanks{Manuscript received XXX; revised XXX.}}

\markboth{To appear in IEEE TRANSACTIONS ON PATTERN ANALYSIS AND MACHINE INTELLIGENCE, 2018}%
{Zhou \MakeLowercase{\textit{et al.}}: Bare Advanced Demo of IEEEtran.cls for Journals}

\IEEEtitleabstractindextext{%


\begin{abstract}

Recovering 3D full-body human pose is a challenging problem with many applications. It has been successfully addressed by motion capture systems with body worn markers and multiple cameras. In this paper, we address the more challenging case of not only using a single camera but also not leveraging markers: going directly from 2D appearance to 3D geometry. Deep learning approaches have shown remarkable abilities to discriminatively learn 2D appearance features. The missing piece is how to integrate 2D, 3D and temporal information to recover 3D geometry and account for the uncertainties arising from the discriminative model. We introduce a novel approach that treats 2D joint locations as latent variables whose uncertainty distributions are given by a deep fully convolutional neural network. The unknown 3D poses are modeled by a sparse representation and the 3D parameter estimates are realized via an Expectation-Maximization algorithm, where it is shown that the 2D joint location uncertainties can be conveniently marginalized out during inference. Extensive evaluation on benchmark datasets shows that the proposed approach achieves greater accuracy over state-of-the-art baselines. Notably, the proposed approach does not require synchronized 2D-3D data for training and is applicable to ``in-the-wild'' images, which is demonstrated with the MPII dataset. 

\end{abstract}

}

\maketitle
\section{Introduction}

This paper is concerned with the challenge of recovering the 3D full-body human pose from a markerless monocular RGB image sequence.
Potential applications of our work include human-computer interaction, surveillance, rehabilitation, sports, video browsing and indexing, and virtual reality.
Typical solutions for this task include motion capture (MoCap) systems with multiple cameras and reflective markers and depth sensors, e.g., Microsoft Kinect \cite{kinect}. These techniques require customized devices, are limited to applications in constrained environments, and cannot be applied to archival RGB images or videos. This paper addresses the pose recovery challenge by using a single camera and avoiding the use of markers: 
going directly from 2D appearance to 3D geometry.

From a geometric perspective, 3D articulated pose recovery is inherently ambiguous from monocular imagery \cite{lee1985determination}. A considerable amount of work 
has tackled the geometric problem to reconstruct 3D human pose from 2D correspondences via articulated constraints \cite{taylor2000reconstruction}, low-rank priors \cite{bregler2000recovering}, sparse representations \cite{ramakrishna2012reconstructing}, or tracking with a body model \cite{bregler1998tracking}. These approaches typically assume 2D correspondences are provided or require careful initialization for frame-to-frame tracking based on low-level image features. Finding 2D correspondences is rendered difficult due to the large variation in human appearance (e.g., clothing, body shape, and illumination), arbitrary camera viewpoint, and obstructed visibility due to  self-occlusions and external entities. Notable successes in 2D pose estimation have used 
discriminatively trained 2D part models coupled with 2D deformation priors, e.g., \cite{yang2011articulated,andriluka20142d,nie2015joint}, and more recently using deep learning, e.g., \cite{toshev2014deep}. Here, the 3D pose geometry is not leveraged. Combining robust image-driven 2D part detectors, expressive 3D geometric pose priors and temporal models to aggregate information over time is a promising area of research that has been given limited attention, e.g., \cite{andriluka2010monocular,zhou2014spatio}.  The challenge posed is how to seamlessly integrate 2D, 3D and temporal information to fully account for the model and measurement uncertainties.

\begin{figure}[t]
   \includegraphics[width=\linewidth]{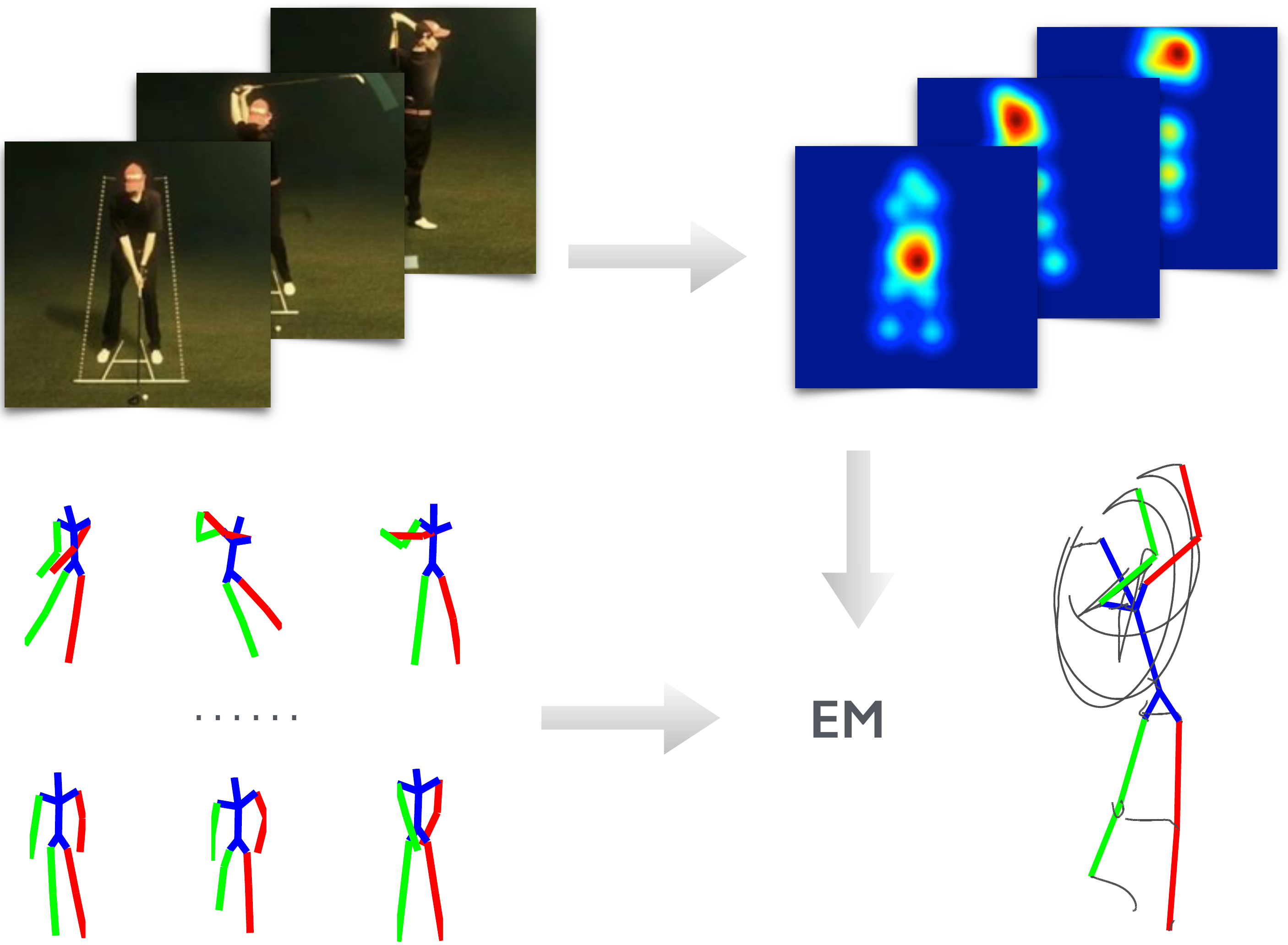}
   \caption{Overview of  proposed approach. 
   (top-left) Input image sequence, (top-right) CNN-based  heat map outputs
   representing the soft localization of 2D joints,
   (bottom-left) 3D pose dictionary, and (bottom-right) the recovered 3D pose sequence reconstruction. To fully account for uncertainty, the problem is addressed in a probabilistic framework where the 2D joint locations are modeled as latent variables and marginalized in an EM algorithm. Temporal smoothness in 3D is also imposed.}
   \label{fig:overview}
   \vspace{-1em}
\end{figure}

This paper presents a 3D human pose estimation framework called MonoCap that consists of a novel synthesis between
discriminative image-based and 3D reconstruction approaches.
In particular, the approach reasons jointly about image-based 2D part location estimates and model-based 3D pose reconstruction, so that they can benefit from each other.  Further,
to improve the approach's robustness against detector error, occlusion, and reconstruction ambiguity, temporal smoothness is imposed on the 3D pose and viewpoint parameters.
\refFig{fig:overview} provides an overview of our approach.
Given the input video (Fig.\ \ref{fig:overview}, top-left), 2D joint heat maps capturing positional uncertainty are generated with a deep fully convolutional neural network (CNN)  (Fig.\ \ref{fig:overview}, top-right).  These heat maps are combined with a sparse model of 3D human pose (Fig.\ \ref{fig:overview}, bottom-left) within an Expectation-Maximization (EM) framework to recover the 3D pose sequence (Fig.\ \ref{fig:overview}, bottom-right).

\subsection{Related work}

Considerable research has addressed the challenge of human motion capture from imagery \cite{moeslund2006survey,sminchisescu20073d,brubaker2010video,deva2011_Book,sarafianos20163d}.
This work includes 2D human pose recovery in both single images (e.g., \cite{yang2011articulated,toshev2014deep,chen2014articulated,jain2014learning,tompson2014joint}), and video, e.g., \cite{sapp2011,cherian2014,nie2015joint,park2015articulated,pfister2015flowing,zhang2015human,newell2016stacked}.
In the current work, focus is placed on 3D pose recovery, where the pose model and prior are expressed in their natural 3D domain.

Early research on 3D monocular pose estimation in videos largely centered on generative models for frame-to-frame pose tracking, e.g.,  \cite{bregler1998tracking,sminchisescu2003kinematic}. 
These approaches rely on a given pose and dynamic model to constrain the pose search space.
Notable drawbacks of this approach include: the requirement that the initialization be provided, and their inability to recover from tracking failures. To address these limitations, bottom-up models were proposed in more recent works, e.g., the ``loose-limbed people'' \cite{sigal2012loose} and ``tracking-by-detection'' \cite{andriluka2010monocular}.

Another strand of research has focused on discriminative methods that predict 3D poses by searching a database of exemplars \cite{shakhnarovich2003fast,mori2006recovering,jiang20103d,yasin2016dual} or via a discriminatively learned mapping from the image directly to human joint locations \cite{agarwal2006recovering,bo2010twin,salzmann2010implicitly,yu2013unconstrained,ionescu2014human,kostrikov2014depth}.
Recently, deep convolutional networks (CNNs) have emerged as a common element behind many state-of-the-art approaches, including 3D human pose estimation, e.g., \cite{li20143d,li2015maximum,tekin2015predicting,du2016marker,park20163d,zhou2016deep}. To deal with the scarcity of training data, some recent works synthesize training images via graphics rendering \cite{chen2016synthesizing} or image mosaicing \cite{rogez2016mocap}.     

Most closely related to our work are generic factorization approaches for recovering 3D non-rigid shapes from image sequences captured with a single camera \cite{bregler2000recovering,akhter2011trajectory,dai2012simple,zhu2014complex,cho2015complex}, i.e., non-rigid structure from motion (NRSFM),
and human pose recovery models based on known skeletons \cite{lee1985determination,taylor2000reconstruction,valmadre2010deterministic,park20113d,radwan2013monocular,leonardos2016articulated} or sparse representations \cite{ramakrishna2012reconstructing,fan2014pose,akhter2015pose,zhou20153d,zhou2015sparse}. Much of this work has been realized by assuming manually labeled 2D joint locations; however, there is some recent work that has used a 2D pose detector to automatically provide the input joints \cite{simo2012single,wang2014robust} or solved 2D and 3D pose estimation jointly \cite{simo2013joint,zhou2014spatio}.

\subsection{Contributions}

In the light of previous work, the proposed approach advances the state-of-the-art in the following ways.
First, in contrast to prediction approaches (e.g., \cite{li2015maximum,tekin2015predicting}), our approach does not require synchronized 2D-3D data, as captured by MoCap systems. The proposed approach only requires readily available annotated 2D imagery (e.g., the ``in-the-wild'' MPII dataset \cite{andriluka14cvpr}) to train a CNN part detector and a separate 3D MoCap dataset (e.g., the CMU MoCap database) for the pose dictionary. The flexibility of using separate sources of training data makes the proposed approach more widely applicable.
In comparison to examplar-based methods (e.g., \cite{jiang20103d,yasin2016dual}), the proposed approach does not need to store and enumerate all possible 2D views and can generalize to unseen poses. 
Compared to other 3D reconstruction methods (e.g., \cite{ramakrishna2012reconstructing,zhou2015sparse}), the proposed approach does not rely on a hard decision of 2D correspondences before reconstruction and considers an arbitrary pose uncertainty. 
In contrast to prior work that consider model-image alignment (e.g., \cite{guan2009estimating,sigal2012loose,bogo2016keep}), the current approach leverages CNNs to learn better 2D representations and sparsity-driven 3D pose optimization to allow efficient and global inference.
Finally, empirical evaluation demonstrates that the proposed approach is more accurate compared to extant approaches.
In particular, in the case where 2D joint locations are provided, the proposed approach exceeds the accuracy of the state-of-the-art NRSFM baseline \cite{dai2012simple} on the Human3.6M dataset \cite{ionescu2014human}. In the case where the 2D joints are unknown, empirical results on the HumanEva I \cite{sigal2010humaneva}, Human3.6M \cite{ionescu2014human}, and KTH Football II \cite{kazemi2013multi} datasets demonstrate overall improvement over published results. Further, the qualitative results on the MPII dataset \cite{andriluka14cvpr} demonstrate that the proposed approach is able to reconstruct 3D poses from single ``in-the-wild'' images with a 3D pose prior learned from a separate MoCap dataset. 

A preliminary version of this work appeared in CVPR 2016 \cite{zhou2016sparseness}. Here, the work is extended in the following ways:
the proposed approach integrates a perspective camera model (as opposed to an orthographic model), a corresponding optimization algorithm is introduced, 
a state-of-the-art 2D pose detector is used, and the empirical evaluations are significantly expanded to make them more comprehensive.
The code is available at \url{https://github.com/daniilidis-group/monocap}.

\section{Models}

In this section, the models that describe the relationships between 3D poses, 2D poses, and images are introduced.

\subsection{Sparse representation of 3D poses}

The 3D human pose is represented by the 3D locations of a set of $p$ joints, denoted by $\bfS_t\in\RR{3}{p}$ for frame $t$.
In general, single-view reconstruction is an ill-posed problem. To address this problem, it is assumed that a 3D pose can be represented as a linear combination of predefined basis poses:
\begin{align}\label{eq:shape-model}
    \bfS_t = \sum_{i=1}^{k} c_{it}\bfB_i,
\end{align}
where $\bfB_i\in\RR{3}{p}$ denotes a basis pose and $c_{it}$ the corresponding weight. The basis poses are learned from training poses provided by a MoCap dataset.
Instead of using the conventional active shape model \cite{cootes1995}, where the basis set is relatively small, a sparse representation is adopted which has been shown in recent work to be
capable of modelling the large variability of human pose, e.g., \cite{ramakrishna2012reconstructing,akhter2015pose,zhou2015sparse}.
That is, an overcomplete dictionary, $\{\bfB_1,\cdots,\bfB_k\}$, is learned with a relatively large number of basis poses, $k$, where
the coefficients, $c_{it}$, are assumed to be sparse. In the remainder of this paper,
$\bfc_t = [c_{1t},\cdots,c_{kt}]^\top$ denotes the coefficient vector for frame $t$ and $\bfC$ the matrix composed of all $\bfc_{t}$.

\subsection{Dependence between 2D and 3D poses}

\subsubsection{Orthographic projection model}

When the camera intrinsic parameters are unknown, an orthographic camera model is used to describe the dependence between a 3D pose and its imaged 2D pose:
\begin{align}\label{eq:camera-model}
    \bfW_t = \bfR_t\bfS_t + \bfT_t\bfone^\top,
\end{align}
where $\bfW_t\in\RR{2}{p}$ denotes the 2D pose in frame $t$, and $\bfR_t\in\RR{2}{3}$ and $\bfT_t\in\R{2}$ the camera rotation and translation, respectively. In the following, $\bfW$, $\bfR$ and $\bfT$ denote the collections of $\bfW_t$, $\bfR_t$ and $\bfT_t$ for all $t$, respectively.

Considering the observation noise and model error, the conditional distribution of the 2D poses given the 3D pose parameters is modeled as
\begin{align}\label{eq:likelihood}
\pr(\bfW|\theta) \propto e^{-\mathcal{L}(\theta;\bfW)},
\end{align}
where $\theta=\{\bfC,\bfR,\bfT\}$ is the union of all the 3D pose parameters
and the loss function, $\mathcal{L}(\theta;\bfW)$, is defined as
\begin{align}\label{eq:loss}
\mathcal{L}(\theta;\bfW) = \frac{\nu}{2}\sum_{t=1}^{n}\left\|\bfW_t - \bfR_t\sum_{i=1}^{k} c_{it}\bfB_i - \bfT_t\bfone^\top\right\|_F^2,
\end{align}
with $\|\cdot\|_F$ denoting the Frobenius norm. The model in \refEq{eq:likelihood} states that given the 3D poses and camera parameters, the 2D location of each joint belongs to a Gaussian distribution with a mean equal to the projection of its 3D counterpart and a precision (i.e., the inverse variance) equal to $\nu$.

The loss function in \refEq{eq:loss} is equivalent to the one proposed in previous work \cite{zhou2015sparse} summed up over frames.  In contrast to previous work \cite{zhou2015sparse}, 
we extend the model to the case of a perspective camera, treat 2D poses as latent variables instead of fixed input, and impose temporal smoothness constraints, as introduced in the following subsections. 

\subsubsection{Perspective camera model}

When the camera intrinsic parameters are given, denoted by the calibration matrix $\bf K$, the perspective camera model is used to describe the dependence between 2D and 3D poses:
\begin{align}\label{eq:model-fp}
    \bfK^{-1}\bfW_t\bfZ_t = \bfR_t\bfS_t + \bfT_t\bfone^\top,
\end{align}
where $\bfW_t\in\RR{3}{p}$ denotes the homogeneous coordinates of 2D joints, $\bfR_t\in\RR{3}{3}$ and $\bfT_t\in\R{3}$ are the rotation and translation in 3D, and $\bfZ_t\in\RR{p}{p}$ is a diagonal matrix with diagonal elements denoting the depth values of joints. Correspondingly, the loss function in the perspective case is defined as:
\begin{align}\label{eq:loss-fp}\small
\mathcal{L}(\theta;\bfW) = \frac{\nu}{2}\sum_{t=1}^{n}\left\|\bfK^{-1}\bfW_t\bfZ_t - \bfR_t\sum_{i=1}^{k} c_{it}\bfB_i - \bfT_t\bfone^\top\right\|_F^2,
\end{align}
where $\theta=\{\bfC,\bfR,\bfT,\bfZ\}$.

Note that minimizing the loss in \refEq{eq:loss-fp} yields a trivial solution where all variables converge to zero due to the inherent 
scale ambiguity in the perspective model. To avoid such a trivial solution, the depth of the root joint is enforced to be one by adding the following constraint during optimization:
\begin{align}
z_{1t}=1,
\end{align}
where $z_{1t}$ denotes the first diagonal element of $\bfZ_t$ corresponding to the root joint.

\subsection{Dependence between pose and image}

When 2D poses are given, we assume that the distribution of 3D pose parameters is conditionally independent of the image data. Therefore, the likelihood function of $\theta$ can be factorized as
\begin{align}\label{eq:likelihood-complete}
\pr(\bfI,\bfW|\theta) = \pr(\bfI|\bfW)\pr(\bfW|\theta),
\end{align}
where $\bfI=\{\bfI_1,\cdots,\bfI_n\}$ denotes the image data. The conditional distribution $\pr(\bfW|\theta)$ is given in \refEq{eq:likelihood}. We assume that the likelihood function $\pr(\bfI|\bfW)$ given the image data can be learned discriminatively using a CNN and written as
\begin{align}\label{eq:pr-I-W}
\pr(\bfI|\bfW) \propto \Pi_{t}\Pi_{j}h_j(\bfw_{jt};\bfI_t),
\end{align}
where $\bfw_{jt}$ denotes the image location of joint $j$ in frame $t$, and $h_j(\cdot;\bfI_t)$ represents a mapping from an image $\bfI_t$ to the likelihood of the joint location (termed heat map).
For each joint $j$, the mapping $h_j$ is approximated by a CNN (described in \refSec{sec:cnn}).

\subsection{Prior on model parameters}

The following penalty function on the model parameters is introduced:
\begin{align}\label{eq:prior}
\mathcal{R}(\theta) = \alpha\|\bfC\|_1 + \frac{\beta}{2}\|\nabla_t\bfC\|_F^2 + \frac{\gamma}{2}\|\nabla_t\bfR\|_F^2,
\end{align}
where $\|\cdot\|_1$ denotes the $\ell_1$-norm (i.e., the sum of absolute values), $\nabla_t$ the discrete temporal derivative operator, and
$\alpha, \beta, \gamma$ are scalar weights. 
The first term penalizes the cardinality of the pose coefficients to induce a sparse pose representation. The second and third terms impose first-order smoothness constraints on both the pose coefficients and rotations. 
A similar smoothness constraint could be imposed on the translation component; however, empirically we did not observe an obvious performance difference with its inclusion.

\section{3D pose inference}

In this section, our approach to 3D pose inference is described.  Here, two cases are distinguished: (i) the image locations of the joints are provided (Section\ \ref{sec:known}), and
(ii) the joint locations are unknown (Section \ref{sec:unknown}).

\subsection{Given 2D poses}\label{sec:known}

When the 2D poses, $\bfW$, are given, the model parameters, $\theta$, are recovered via penalized maximum likelihood estimation (MLE):
\begin{align}\label{eq:pmle}
\theta^* &= \argmax_{\theta} ~ \ln\pr(\bfW|\theta) - \mathcal{R}(\theta) \nonumber \\
& = \argmin_{\theta} ~ \mathcal{L}(\theta;\bfW) + \mathcal{R}(\theta).
\end{align}
The problem in \refEq{eq:pmle} is solved via block coordinate descent, i.e., alternately updating one block of variables while fixing the others. 

\subsubsection{Orthographic projection model}

Under the orthographic model, the loss function is given in \refEq{eq:loss} and the update rules in each iteration are given below. 

The update of $\bfC$ needs to solve:
\begin{align}\label{eq:C-step}
\bfC \leftarrow \argmin_{\bfC} ~~ \mathcal{L}(\bfC;\bfW) + \alpha\|\bfC\|_1 + \frac{\beta}{2}\|\nabla_t\bfC\|_F^2,
\end{align}
where the objective is the composite of two differentiable functions plus an $\ell_1$ penalty. The problem in \refEq{eq:C-step} is solved by accelerated proximal gradient (APG) \cite{nesterov2007gradient}.
Since the problem in \refEq{eq:C-step}  is convex, global optimality is guaranteed. 

The update of $\bfR$ needs to solve:
\begin{align}\label{eq:R-step}
\bfR \leftarrow \argmin_{\bfR} ~~ \mathcal{L}(\bfR;\bfW) + \frac{\gamma}{2}\|\nabla_t\bfR\|_F^2,
\end{align}
where the objective is differentiable and the variables are rotations restricted to $SO(3)$. Here, manifold optimization is adopted to update the rotations
using the trust-region solver in the Manopt toolbox \cite{boumal2014manopt}.

The update of $\bfT$ has the following closed-form solution:
\begin{align}\label{eq:T-step}
\bfT_t \leftarrow \mbox{row mean}\left\{\bfW_t - \bfR_t\bfS_t\right\}.
\end{align}

\subsubsection{Perspective camera model}

Under the perspective model, the loss function is given in \refEq{eq:loss-fp} and the update rules in each iteration are given below. 

The update rules for $\bfC$ and $\bfR$ have the same forms as the orthographic case given in \refEq{eq:C-step} and \refEq{eq:R-step}, respectively.

The update of $\bfT$ has the following closed-form solution:
\begin{align}\label{eq:T-step-fp}
\bfT_t \leftarrow \mbox{row mean}\left\{\bfK^{-1}\bfW_t\bfZ_t - \bfR_t\bfS_t\right\}.
\end{align}

The update of $\bfZ$ is also closed-form. Suppose $z_{it}$ denotes the $i$-th diagonal element of $\bfZ_t$, $\bfu_{it}$ the $i$-th column of $\bfK^{-1}\bfW_t$, and $\bfv_{it}$ the $i$-th column of $\bfR_t\bfS_t+\bfT_t\bfone^\top$. Then the solution for each $z_{it}$ is  
\begin{align}\label{eq:Z-step}
z_{it}\leftarrow 
\begin{cases}
     1 & \mbox{if $i=1$}, \\
    \frac{\bfu_{it}^\top\bfv_{it}}{\bfu_{it}^\top\bfu_{it}} & \mbox{otherwise}
\end{cases}.
\end{align}

The entire algorithm for 3D pose inference given the 2D poses is summarized in \refAlg{alg:bcd}.
The iterations are terminated once the objective value has converged.
Since in each step the objective function is non-increasing, the algorithm is guaranteed to terminate; however, since the problem in \refEq{eq:pmle} is nonconvex, the algorithm requires a suitably chosen initialization (described in \refSec{sec:initialization}).

\begin{algorithm}[t]\label{alg:bcd}
\LinesNumbered
\caption{Block coordinate descent to solve \refEq{eq:pmle} under the orthographic or perspective projection camera models.}
\vspace{0.3em}
\KwIn{$\bfW$ \tcp*[r]{\small 2D joint locations}}
\KwOut{$\theta$ \tcp*[r]{\small 3D pose parameters}}
\vspace{0.3em}
initialize the parameters \tcp*{\small \refSec{sec:initialization}}
\While{not converged}{
update $\bfC$ by \refEq{eq:C-step} with APG \;
update $\bfR$ by \refEq{eq:R-step} with Manopt \;
update $\bfT$ by \refEq{eq:T-step} if using orthographic model \\
~~~~~~~~~~~~or by \refEq{eq:T-step-fp} if using perspective model \;
update $\bfZ$ by \refEq{eq:Z-step} if using perspective model \;
}
\end{algorithm}

\subsection{Unknown 2D poses}\label{sec:unknown}

If the 2D poses are unknown, $\bfW$ is treated as a latent variable and is marginalized out during the estimation process. The marginalized likelihood function is
\begin{align}
\pr(\bfI|\theta) = \int\pr(\bfI,\bfW|\theta)d\bfW, \label{eq:marginalization}
\end{align}
where $\pr(\bfI,\bfW|\theta)$ is given in \refEq{eq:likelihood-complete}.

Direct marginalization of (\ref{eq:marginalization}) is intractable.
Instead, an EM algorithm is developed to compute the penalized MLE.
In the expectation step, the expectation of the penalized log-likelihood is calculated with respect to the conditional distribution of $\bfW$ given the image data and the previous estimate of all the 3D pose parameters, $\theta'$:
\begin{align}
&Q(\theta|\theta') = \int \left\{\ln\pr(\bfI,\bfW|\theta) - \mathcal{R}(\theta)\right\} ~ \pr(\bfW|\bfI,\theta') d\bfW \nonumber \\
&= \int \left\{\ln\pr(\bfI|\bfW) + \ln\pr(\bfW|\theta) - \mathcal{R}(\theta)\right\} \pr(\bfW|\bfI,\theta') d\bfW \nonumber \\
&= \mbox{const} - \int \mathcal{L}(\theta;\bfW) \pr(\bfW|\bfI,\theta') d\bfW - \mathcal{R}(\theta).
\end{align}

It can be shown that (see Appendix for the derivation)
\begin{align}\label{eq:der1}
\int \mathcal{L}(\theta;\bfW) \pr(\bfW|\bfI,\theta') d\bfW = \mathcal{L}(\theta;\mathbb{E}\left[\bfW|\bfI,\theta'\right]) + \mbox{const},
\end{align}
where $\mathbb{E}\left[\bfW|\bfI,\theta'\right]$ is the expectation of $\bfW$ given $\bfI$ and $\theta'$:
\begin{align}\label{eq:der2}
\mathbb{E}\left[\bfW|\bfI,\theta'\right]
&= \int \pr(\bfW|\bfI,\theta')~ \bfW ~ d\bfW \nonumber \\
&= \int \frac{\pr(\bfI|\bfW)\pr(\bfW|\theta')}{M} ~ \bfW ~ d\bfW,
\end{align}
and $M$ is a scalar that normalizes the probability. 
Both $\pr(\bfI|\bfW)$ and $\pr(\bfW|\theta')$ given in \refEq{eq:pr-I-W} and \refEq{eq:likelihood}, respectively, are products of marginal probabilities of $\bfw_{jt}$. Therefore,
the expectation of each $\bfw_{jt}$ can be computed separately. In particular, the expectation of each $\bfw_{jt}$ is efficiently approximated by sampling over the pixel grid.

In the maximization step, the following is computed:
\begin{align}
\theta \leftarrow &\argmax_{\theta} Q(\theta|\theta') \nonumber \\
= &\argmin_{\theta} ~~ \mathcal{L}(\theta;\mathbb{E}\left[\bfW|\bfI,\theta'\right]) + \mathcal{R}(\theta),
\end{align}
which can be solved by \refAlg{alg:bcd}.

The entire EM algorithm is summarized in \refAlg{alg:em} with the initialization scheme described next in \refSec{sec:initialization}.

\begin{algorithm}[t]\label{alg:em}
\LinesNumbered
\caption{The EM algorithm for pose from video.}
\vspace{0.3em}
\KwIn{$h_j(\cdot;\bfI_t),~ \forall j,t$ \tcp*[r]{\small heat maps}}
\KwOut{$\theta=\{\bfC,\bfR,\bfT,\bfZ\}$ \tcp*[r]{\small pose parameters}}
\vspace{0.3em}
initialize the parameters \tcp*[r]{\small \refSec{sec:initialization}}
\While{not converged}{
$\theta'=\theta$\;
\vspace{0.3em}
\tcp{\small Compute the expectation of $\bfW$}
$\mathbb{E}\left[\bfW|\bfI,\theta'\right] = \int \frac{1}{M}\pr(\bfI|\bfW)\pr(\bfW|\theta') ~ \bfW ~ d\bfW$\;
\vspace{0.3em}
\tcp{\small Update $\theta$ by \refAlg{alg:bcd}}
$\theta = \argmin_{\theta} ~~ \mathcal{L}(\theta;\mathbb{E}\left[\bfW|\bfI,\theta'\right]) + \mathcal{R}(\theta)$ \;
}
\vspace{0.5em}
\end{algorithm}

\subsection{Initialization and dictionary learning}\label{sec:initialization}

A convex relaxation approach \cite{zhou20153d,zhou2015sparse} is used to initialize the parameters.
This convex formulation was initially proposed for pose recovery in a single image frame given 
2D correspondences \cite{zhou20153d}, a special case of \refEq{eq:pmle}.
The approach was later extended to handle 2D correspondence outliers \cite{zhou2015sparse}. If the 2D poses are given, the model parameters are initialized for each frame separately with the convex relaxation \cite{zhou2015sparse}.
Alternatively, if the 2D poses are unknown, for each joint, the image location with the maximum heat map value is used.
Next, the robust estimation algorithm from \cite{zhou2015sparse} is applied to initialize the parameters.

A dictionary learning algorithm \cite{zhou2015sparse} is used to learn the pose dictionaries given training pose data. The dictionary size is empirically set to $K=64$ for action specific dictionaries and $K=128$ for the non-action specific case, based on the trade-off between reconstruction error and computational efficiency.

\section{CNN-based joint uncertainty regression}\label{sec:cnn}

\begin{figure}
  \centering
  \includegraphics[width=\linewidth]{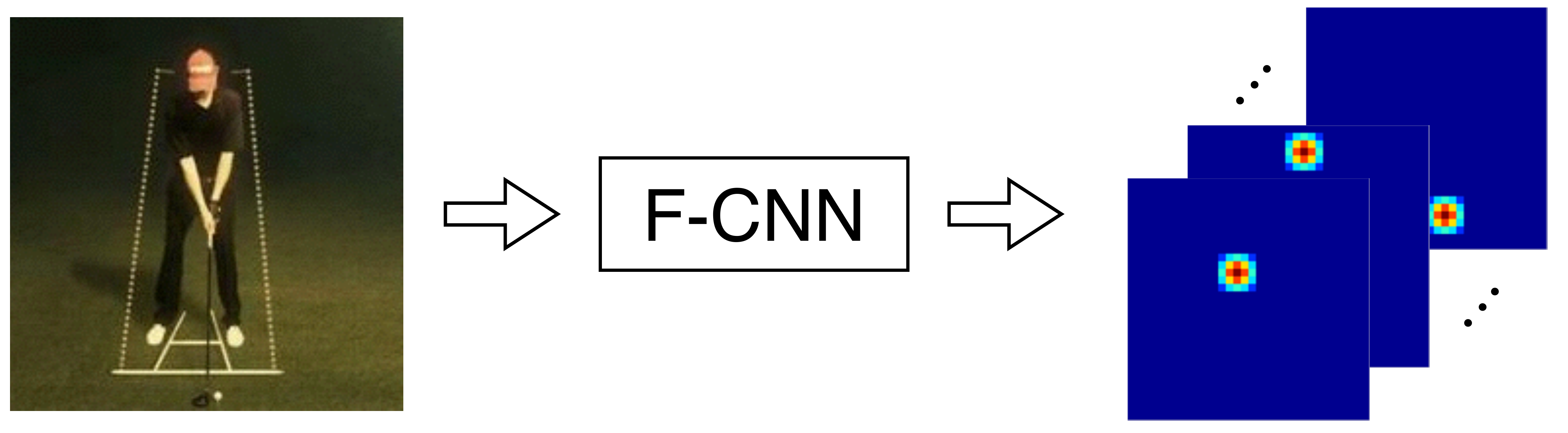}
  \caption{CNN-based 2D joint regressor. The network is a fully convolutional neural network (F-CNN). The input is an image and the output is a multi-channel heat map with each channel capturing the spatial uncertainty distribution of a joint. }\label{fig:cnn}
\end{figure}

A CNN is used to learn the mapping $\bfI_t \mapsto h_j(\cdot;\bfI_t)$, where $\bfI_t$ denotes an input image and $h_j(\cdot;\bfI_t)$ a (spatial) heat map for joint $j$. Rather than training $p$ separate networks for $p$ joints, a fully convolutional neural network \cite{long2015fully} is trained to regress $p$ joint distributions simultaneously by taking into account the full-body information. 
The training labels to be regressed are multi-channel heat maps with each channel corresponding to the image location uncertainty distribution for a joint. The uncertainty is modeled by a Gaussian centered at the annotated joint location. \refFig{fig:cnn} illustrates the CNN-based 2D joint regressor.

The Stacked Hourglass model proposed by Newell et al. \cite{newell2016stacked} is adopted as the network architecture, which represents the
state-of-the-art for 2D human pose detection. The network is fully convolutional and the shape of the network is an hourglass structure consisting of a series of downsampling layers with decreasing resolutions followed by a series of upsampling layers.  This implements bottom-up and top-down processing to integrate contextual information over the entire image. A second hourglass component is stacked at the end of the first one to refine the initial heat maps. 
The final outputs are $64 \times 64$ heat maps. The $\ell_2$ loss is minimized during training and intermediate supervision is applied at the end of the first module. The convolutional layers are implemented with residual modules. Please refer to the original paper \cite{newell2016stacked} for details. 

During testing, consistent with previous 3D pose methods (e.g., \cite{li2015maximum,tekin2015predicting}), a bounding box around the subject is assumed.
The image patch in the bounding box, $\bfI_t$, is cropped in frame $t$ and is provided to the network as input to predict the heat maps, $h_j(\cdot;\bfI_t),~\forall j=1,\dotsc,n$.

\section{Empirical evaluation}

\subsection{Datasets}

\begin{table*}
\caption{Mean per joint errors (mm) on Human3.6M \cite{ionescu2014human} given 2D joints.}\label{tab:hm36m_rec}
\centering
\renewcommand{\arraystretch}{1.5}
\begin{tabular}{l*{15}{c}}
\toprule
 & Directions & Discussion & Eating & Greeting & Phoning & Photo & Posing & Purchases \\
\toprule
PMP  \cite{ramakrishna2012reconstructing}  &     127.6 &        134.8 &        127.6 &        144.7 &        129.7 &        138.4 &        137.0 &        146.2 \\
NRSFM \cite{dai2012simple}  &  137.2 &        140.7 &        129.8 &        118.0 &        134.3 &        120.0 &        156.5 &        161.7 \\
Convex \cite{zhou2015sparse} & 94.8 &         89.7 &         83.3 &         99.1 &         85.2 &        109.2 &         95.3 &         84.9 \\
Orth.  & 92.8 &         88.2 &         82.3 &         97.4 &         83.3 &        107.1 &         93.2 &         83.7         \\
Persp.  & 73.5 &         68.0 &         81.5 &         77.5 &         71.0 &         90.8 &         72.9 &         77.3         \\
\toprule
 & Sitting & SittingDown & Smoking & Waiting & WalkDog & Walking & WalkTogether & Average \\
 \toprule
PMP  \cite{ramakrishna2012reconstructing}  &  149.8 &        161.5 &        132.0 &        156.7 &        120.2 &        159.3 &        159.8 &        141.4 \\
NRSFM \cite{dai2012simple} &  180.6 &        177.2 &        127.2 &        137.2 &        136.8 &        104.7 &        113.6 &        136.9 \\
Convex \cite{zhou2015sparse} & 80.3 &         97.2 &         82.2 &         94.4 &         83.4 &         81.4 &         92.3 &         90.2 \\
Orth. & 79.0 &         96.8 &         80.9 &         92.4 &         81.6 &         81.5 &         91.3 &         88.7 \\
Persp.  & 68.3 &         91.4 &         66.8 &         75.0 &         67.0 &         60.1 &         71.5 &         74.2 \\
\toprule
\end{tabular}
\end{table*}

\begin{table}
\caption{Mean reconstruction errors (mm) on Human3.6M \cite{ionescu2014human} given 2D joints.}\label{tab:nrsfm}
\centering
\renewcommand{\arraystretch}{1.5}
\begin{tabular}{l*{15}{c}}
\toprule
 & \cite{ramakrishna2012reconstructing} &  \cite{dai2012simple} & \cite{zhou2015sparse} & \cite{yasin2016dual} & Orth. & Persp. \\
\toprule
Protocol I & 90.9 & 82.9 & 51.3 & - & 51.1 & 50.5 \\
Protocol II & - & - & - & 70.5 & 55.1 & 54.6 \\
\toprule
\end{tabular}
\end{table}

Empirical evaluation was performed on four standard datasets -- Human3.6M \cite{ionescu2014human}, Human Eva I \cite{sigal2010humaneva}, KTH Football II \cite{kazemi2013multi}, and MPII \cite{andriluka14cvpr}.  These datasets cover both controlled lab and more realistic scenarios. The first three datasets were used for quantitative evaluation and the last one for qualitative evaluation.

\subsection{Evaluation metric} 
\label{sub:evaluation_metric}

Given a set of estimated 3D joint locations $\hat\bfx_1,\cdots,\hat\bfx_n$ and the corresponding ground-truth locations $\bfx_1^*,\cdots,\bfx_n^*$ in the same coordinates, the \textbf{per joint error} is defined as the average Euclidean distance over all the joints:
\begin{align}\label{eq:pje}
e = \frac{1}{n}\sum_{i=1}^n\|\hat\bfx_i-\bfx_i^*\|_2.
\end{align}
Note that the above metric depends on the absolute pose of the estimated structure, including scale, translation, and orientation. Scale and depth ambiguities are inherent to monocular reconstruction and in general cannot be resolved. The scale is directly learned from training subjects in prediction-based approaches \cite{ionescu2014human,li2015maximum,tekin2015predicting}. For a fair comparison, our reconstruction output is scaled such that the mean limb length is identical to the average value of all training subjects. As the standard protocol in the Human3.6M and HumanEva datasets, the root locations of compared skeletons are aligned to make the evaluation translation invariant. Note that Procrustes alignment to the ground truth is not allowed.

The \textbf{reconstruction error} is defined as the 3D per joint error up to a similarity transformation, $\mathcal{T}$:
\begin{align*}
r = \min_{\mathcal{T}}\frac{1}{n}\sum_{i=1}^n\|\hat\bfx_i-\mathcal{T}(\bfx_i^*)\|_2,
\end{align*}
where 
the optimal parameters are obtained by Procrustes alignment. The 3D reconstruction error is widely used in structure-from-motion to evaluate the accuracy of recovered structure regardless of scale and rigid pose.

The \textbf{percentage of correct parts (PCP)} is defined as
\begin{align}\label{eq:pcp}
\mbox{PCP} = \frac{1}{n}\sum_{i=1}^n \mathbb{I}\left(\frac{\|\hat{\bfx}_i-\bfx_i\|+\|\hat{\bfy}_i-\bfy_i\|}{2\|\bfx_i-\bfy_i\|}\leq \tau\right),
\end{align}
where $\bfx_i$ and $\bfy_i$ are the coordinates of two ends of the $i$-th part and $\hat\bfx_i$ and $\hat\bfy_i$ the corresponding estimates. $\mathbb{I}$ and $\tau$ denote the indicator function and the threshold, respectively. The PCP metric measures the fraction of correctly located parts with respect to a given threshold ($\tau=0.5$ in this work).

\subsection{Human3.6M}
\label{sec:h36m}

\begin{table*}
\caption{Mean per joint errors (mm) on Human3.6M \cite{ionescu2014human}. Marker $\dagger$ indicates using sequence information.}
\centering
\renewcommand{\arraystretch}{1.35}
\begin{tabular}{l*{15}{c}}
\toprule
 & Directions & Discussion & Eating & Greeting & Phoning & Photo & Posing & Purchases \\
\toprule
LinKDE \cite{ionescu2014human} & 132.7 & 183.5 & 132.3 & 164.3 & 162.1 & 205.9 & 150.6 & 171.3 \\
Li et al. \cite{li2015maximum} & - & 136.8 & 96.9 & 124.7 & - & 168.6 & - & - \\
Tekin et al. \cite{tekin2015predicting}$^\dagger$ & 102.4 & 147.7 & 88.8 & 125.2 & 118.0 & 182.7 & 112.3 & 129.1 \\
Du et al. \cite{du2016marker}$^\dagger$  &        85.0 &       112.6 &        104.9 &        122.0 &        139.0 &        135.9 &        105.9 &        166.1 \\
Park et al. \cite{park20163d}  &         100.3 &        116.1 &        89.9 &        116.4 &        115.3 &        149.5 & 117.5 &        106.9 \\
Zhou et al. \cite{zhou2016deep}  &        91.8 &        102.4 &        96.6 &        98.7 &        113.3 &        125.2 &       90.0 &        93.8 \\
Generic+nonspecific$^\dagger$ & 82.8 &         88.2 &         93.3 &         93.0 &        111.7 &        115.9 &         85.4 &        131.4 \\
Generic+specific$^\dagger$ & 71.8 &         84.6 &         85.4 &         84.4 &        106.6 &        120.4 &         81.7 &        128.6 \\
Fine-tuned+nonspecific$^\dagger$ & 77.8 &         74.0 &         85.0 &         83.7 &         80.1 &         96.4 &         79.2 &         85.3 \\
Fine-tuned+specific$^\dagger$ & 69.2 &         75.2 &         75.8 &         73.6 &         75.4 &         99.6 &         76.1 &         73.6 \\
Fine-tuned+specific w/o smooth & 71.4 &         77.0 &         75.7 &         77.2 &         76.6 &        102.3 &         79.3 &         75.0 \\
\toprule
 & Sitting & SittingDown & Smoking & Waiting & WalkDog & Walking & WalkTogether & Average \\
\toprule
LinKDE \cite{ionescu2014human} & 151.5 & 243.0 & 162.1 & 170.6 & 177.1 & 96.6 & 127.8 & 162.1  \\
Li et al. \cite{li2015maximum} & - & - & - & - & 132.1 & 69.9 & - & - \\
Tekin et al. \cite{tekin2015predicting}$^\dagger$ & 138.8 & 224.9 & 118.4 & 138.7 & 126.2 & {55.0 }& {65.7 }& 124.9  \\
Du et al. \cite{du2016marker}$^\dagger$  &        117.4 &       226.9&        120.0 &        117.6 &        137.3 &        99.2 &        106.5 &        126.4 \\
Park et al. \cite{park20163d}  &         137.2 &        190.8 &        105.7 &        125.1 &       131.9 &        62.6 &        96.1 &        117.3 \\
Zhou et al. \cite{zhou2016deep}  &        132.1 &        158.9 &        106.9 &        94.4 &        126.0 &        79.0 &        98.9 &        107.2 \\
Generic+nonspecific$^\dagger$ & 126.8 &        226.8 &         97.6 &         91.7 &         99.7 &         83.5 &         88.4 &        107.8 \\
Generic+specific$^\dagger$ & 114.9 &        225.1 &         93.3 &         99.0 &         95.6 &         65.0 &         74.9 &        102.2 \\
Fine-tuned+nonspecific$^\dagger$ & 79.1 &        118.4 &         75.3 &         80.3 &         75.3 &         67.8 &         81.8 &         82.7 \\
Fine-tuned+specific$^\dagger$ & 75.0 &        109.6 &         73.7 &         88.9 &         71.8 &         55.6 &         73.5 &         77.8 \\
Fine-tuned+specific w/o smooth &         76.0 &        112.2 &         74.2 &         91.3 &         73.1 &         57.8 &         74.1 &         79.6 \\
\toprule
\end{tabular}
\label{tab:h36m}
\end{table*}

\begin{table*}
\caption{Mean {reconstruction errors} (mm) on Human3.6M \cite{ionescu2014human}.}
\centering
\renewcommand{\arraystretch}{1.35}
\begin{tabular}{l*{15}{c}}
\toprule
 & Directions & Discussion & Eating & Greeting & Phoning & Photo & Posing & Purchases \\
\toprule
SMPLify \cite{bogo2016keep} & 62.0 & 60.2 & 67.8 & 76.5 & 92.1 & 77.0 & 73.0 & 75.3 \\
Generic+nonspecific & 52.0 &         54.0 &         59.1 &         61.7 &         74.2 &         70.7 &         51.5 &         60.3 \\
Fine-tuned+nonspecific & 46.7 &         47.7 &         54.9 &         54.1 &         56.3 &         65.4 &         46.9 &         49.1 \\
\toprule
 & Sitting & SittingDown & Smoking & Waiting & WalkDog & Walking & WalkTogether & Average \\
\toprule
SMPLify \cite{bogo2016keep} & 100.3 & 137.3 & 83.4 & 77.3 & 79.7 & 86.8 & 81.7 & 82.3 \\
Generic+nonspecific & 83.9 &        119.9 &         66.9 &         54.8 &         64.5 &         55.6 &         59.1 &         65.9 \\
Fine-tuned+nonspecific & 60.1 &         81.5 &         53.2 &         49.7 &         54.2 &         47.1 &         53.7 &         54.7 \\
\toprule
\end{tabular}
\label{tab:h36m1}
\end{table*}

\begin{table*}
\caption{Mean per joint errors (mm) on the Human3.6M online test set (``H36M\_NOS10'') \cite{ionescu2014human}.}
\centering
\renewcommand{\arraystretch}{1.35}
\begin{tabular}{l*{15}{c}}
\toprule
 & Directions & Discussion &	Eating &	Greeting &	Phoning &	Posing &	Purchases &	Sitting \\
\toprule
Ionescu et al. \cite{ionescu2014human} & 117&	108&	91&	129&	104&	130&	134&	135 \\
Li et al. \cite{li2015maximum} & - & 92& 76& 98& 92& 107& 141& 136 \\
Grinciunaite et al. \cite{grinciunaite2016human} & 91&	89&	94&	102&	105&	99&	112&	151 \\
Proposed & 78   &    74    &   86 &	80   &    89	& 93	& 76    &   95	\\
\toprule
 & SittingDown &	Smoking	& Photo &	Waiting &	Walking &	WalkDog &	WalkTogether &    Mean \\
\toprule
Ionescu et al. \cite{ionescu2014human} & 200&	117&	195&	132&	115&	162&	156&	133  \\
Li et al. \cite{li2015maximum} & 265 & 97& 171& 105& 99& 139& 110& 122 \\
Grinciunaite et al. \cite{grinciunaite2016human} & 239&	109&	151&	106&	101&	141&	106&	119 \\
Proposed & 114  &   83	& 101 &	91 &	64	& 87	& 71 &           86\\
\toprule
\end{tabular}
\label{tab:h36m_test}
\end{table*}

Human3.6M \cite{ionescu2014human} is a recent large-scale dataset for 3D human sensing. It includes millions of 3D human poses acquired from a MoCap system with corresponding images from calibrated cameras.
This setup provides synchronized videos and 2D-3D pose data for evaluation. It includes 11 subjects performing 15 actions, such as eating, sitting, and walking. The same data partition protocol as in previous work was used \cite{li2015maximum,tekin2015predicting}: the data from five subjects (S1, S5, S6, S7, S8) was used for training, and the data from two subjects (S9, S11) was used for testing. The original frame rate is 50 fps and is downsampled to 10 fps.

\subsubsection{3D pose reconstruction with known 2D pose}
 
First, the evaluation of the 3D reconstructability of the proposed approach with known 2D poses is presented. The generic approach to 3D reconstruction from 2D correspondences across a sequence is NRSFM. The proposed approach is compared to the state-of-the-art in NRSFM \cite{dai2012simple} on Human3.6M. A recent approach for single-view pose reconstruction, Projected Matching Pursuit (PMP) \cite{ramakrishna2012reconstructing}, and the initialization approach \cite{zhou2015sparse} used in our pipeline are also included in the comparison.

All sequences of S9 and S11 in Human3.6M were used for evaluation. A single pose dictionary from all the training pose data, irrespective of the action type, was used, i.e., a non-action specific dictionary. Dai et al.'s approach \cite{dai2012simple} requires a predefined rank $K$.  Various values of $K$ were considered with the best result for each sequence reported.

The per joint errors for each action are presented in \refTab{tab:hm36m_rec}, while the reconstruction errors are summarized in \refTab{tab:nrsfm}. In \refTab{tab:nrsfm}, Protocol I means the protocol introduced above in this paper, and Protocol II is the one proposed in \cite{yasin2016dual} where only S11 and 14 joints are used in evaluation. The proposed approach  with a perspective model (``Persp.") outperforms its orthographic counterpart (``Orth.") by a large margin in terms of per joint error.
This performance difference is mostly due to the inaccuracy of rigid pose estimation with the orthographic model, which greatly increases the per joint error as Procrustes alignment is not allowed in the evaluation. Moreover, the orthographic model may suffer from a reflection ambiguity.  
In terms of reconstruction error which ignores rigid pose, the difference between the perspective and orthographic models is much smaller.

``Orth." performs slightly better than the initialization approach \cite{zhou2015sparse}. Note, given the 2D joints, 
``Orth." reduces to the initialization approach 
combined with temporal smoothness constraints.

The proposed approach also outperforms the NRSFM baseline  \cite{dai2012simple}. The main reason is that the videos are captured by stationary cameras. Although the subject is occasionally rotating, the ``baseline" between frames is generally small, and neighboring views provide insufficient geometric constraints for 3D reconstruction. In other words, NRSFM is very difficult to solve in the case of slow camera motion. This observation is consistent with prior findings in the NRSFM literature, e.g., \cite{akhter2011trajectory}. In this case, the structure prior (even a non-action specific one) learned from existing MoCap data is critical for reconstruction.

\subsubsection{3D pose reconstruction with unknown 2D pose}

Next, results on the Human3.6M dataset are reported when 2D poses are predicted from images using the CNN described in \refSec{sec:cnn}. The proposed approach is compared to several recent baselines including both discriminative approaches (e.g., \cite{ionescu2014human,li2015maximum,tekin2015predicting}) and two-stage reconstruction approaches (e.g., \cite{bogo2016keep}). For the proposed approach, four combinations of training data sources were considered -- the generic hourglass model trained on MPII (``generic'') or the fine-tuned model trained on Human3.6M (``fine-tuned'') combined with the nonspecific dictionary learned with all 3D pose data (``nonspecific'') or the dictionary learned with action specific pose data (``specific'').

The mean per joint errors are summarized in \refTab{tab:h36m}. The table shows that the proposed approach outperforms the baselines on most of the actions, yielding a much lower average error compared to the baselines. The ``walk" and ``walk together" sequences involve very predictable and repetitive motions. This might favor the direct regression approach \cite{tekin2015predicting}. The reconstruction errors are provided in \refTab{tab:h36m1}. The proposed approach with a generic 2D pose detector outperforms SMPLify \cite{bogo2016keep}, which is the state-of-the-art two-stage approach that reconstructs 3D poses by fitting a parametric body shape model to 2D joints detected by a CNN-based 2D pose detector.

Comparing the results using different sources of training data, a remarkable improvement was achieved by using the fine-tuned 2D pose detector and the action specific 3D pose dictionaries. Nevertheless, the proposed approach with the generic detector and non-action specific dictionary still attained very competitive performance compared to the state-of-the-art. Note that the proposed approach can also be applied to single frames without temporal smoothness, with corresponding results presented in the last row of \refTab{tab:h36m}. 

\subsubsection{Human3.6M online test set}
\refTab{tab:h36m_test} presents results on the Human3.6M online test set ``H36M\_NOS10''.  The test set includes 360 sequences from three subjects with hidden ground truth. As the standard protocol, all compared approaches used action specific training. 
We used the fine-tuned hourglass model and action specific dictionaries. The results show that the proposed approach achieves significant improvements for all actions compared to the previous approaches.

\subsubsection{Ablative analysis}

\refTab{tab:steps} shows the impact of each component in our approach. For our approach, the 2D pose estimates were obtained from the expectations of their posterior distributions given in \refEq{eq:der2}. Note that the 2D errors are with respect to the normalized bounding box size $256\times256$. Two cases of 2D input were considered: heat maps from the generic detector (less accurate), and the fine-tuned detector (more accurate).  In this evaluation, action specific pose dictionaries and the Stacked Hourglass for joint heat map prediction were used.

\refTab{tab:steps} suggests that, for the same model, both the 3D and 2D errors increase significantly if EM is not applied.  This indicates the importance of marginalizing the 2D uncertainty. Removing the smoothness constraint also increases the error, while the difference is more apparent when the 2D input is more noisy (from the ``generic" detector). The perspective model always outperforms the orthographic model 
with the gap in terms of per joint error being much larger than that of reconstruction error.  This indicates that the performance gain using a perspective camera model is mainly due to the more accurate rigid pose estimation. Finally, with the same orthographic model, the proposed approach clearly improves the initial solution \cite{zhou2015sparse} by taking advantage of EM and smoothness.

An alternative to the proposed EM algorithm is the maximum a posteriori (MAP) estimation which minimizes $-\ln\pr(\bfI,\bfW|\theta)+\mathcal{R}(\theta)$ over $\theta$ and $\bfW$. We implemented the block coordinate descent to solve the MAP estimation. With the fine-tuned detector, the mean per joint error of MAP is 79.3 mm, which is worse than that of EM (77.8 mm).

\begin{table}
\caption{ Estimation errors for the variants of the proposed approach. ``PJ", ``Rec" and ``2D" correspond to per joint error (mm), reconstruction error (mm), and 2D error (pixel), respectively.}
\centering
\renewcommand{\arraystretch}{1.5}
\begin{tabular}{l*{15}{c}}
\toprule
& \multicolumn{3}{c}{Generic} & \multicolumn{3}{c}{Fine-tuned} \\
 & PJ & Rec  & 2D & PJ & Rec & 2D \\
\toprule
Perspective & 102.2 & 63.3 & 10.1 & 77.8 & 53.2 & 6.0 \\
w/o smooth & 106.0 & 65.0 & 10.2 & 79.6 & 54.1 & 6.0 \\
w/o smooth/EM & 108.1 & 67.2 & 11.3 & 81.4 & 55.3 & 6.5 \\
\hline
Orthographic & 118.4 & 64.4 & 10.1 & 92.5 & 55.6 & 6.0 \\
w/o smooth & 122.2 & 65.4 & 10.2 & 93.8 & 55.3 & 6.0 \\
w/o smooth/EM  \cite{zhou2015sparse} & 127.3 & 67.2 & 11.3 & 100.5 & 58.2 & 6.5 \\
\toprule
\end{tabular}
\label{tab:steps}
\end{table}

\subsubsection{Sensitivity to parameters}

\begin{figure}
  \includegraphics[width=0.32\linewidth]{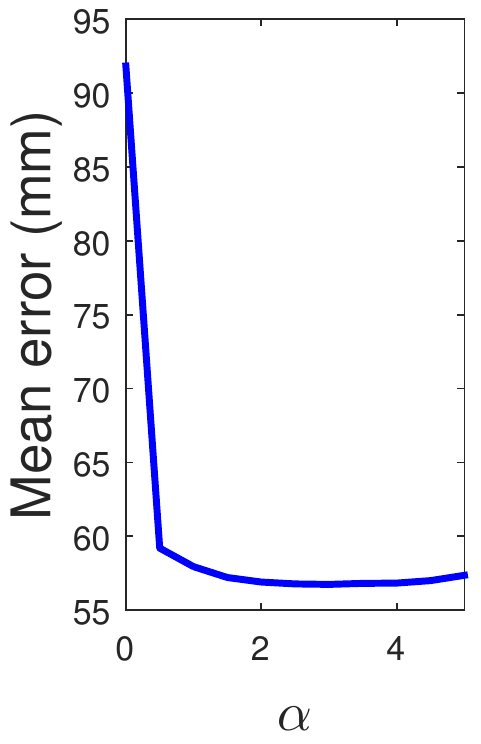} \hfill
  \includegraphics[width=0.32\linewidth]{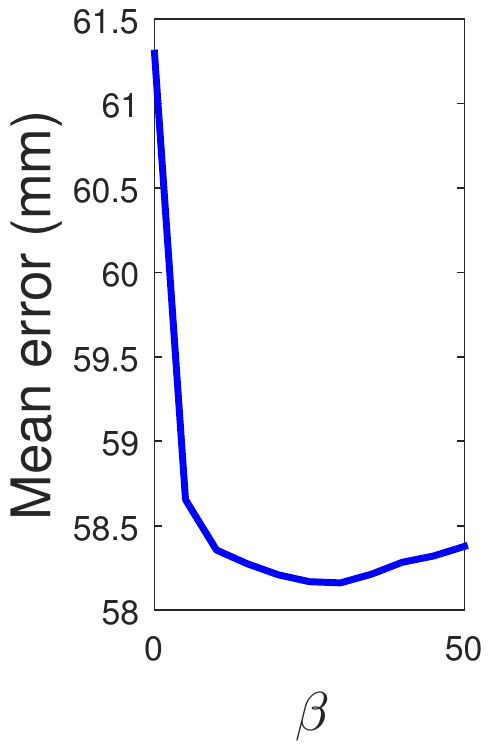} \hfill
  \includegraphics[width=0.32\linewidth]{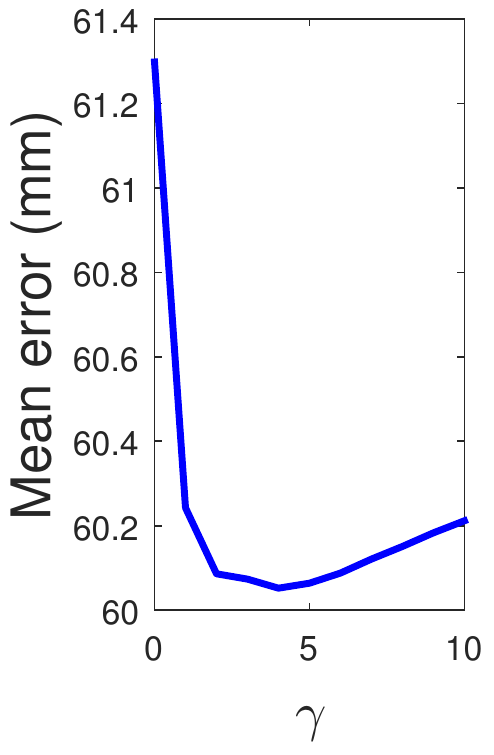}
  \caption{Sensitivity to model parameters. The mean reconstruction error versus model parameter $\alpha$, $\beta$ or $\gamma$ is shown, respectively.}\label{fig:par}
\end{figure}

\begin{figure*}
  \centering
  \begin{minipage}{0.49\textwidth}
  \includegraphics[width=\linewidth]{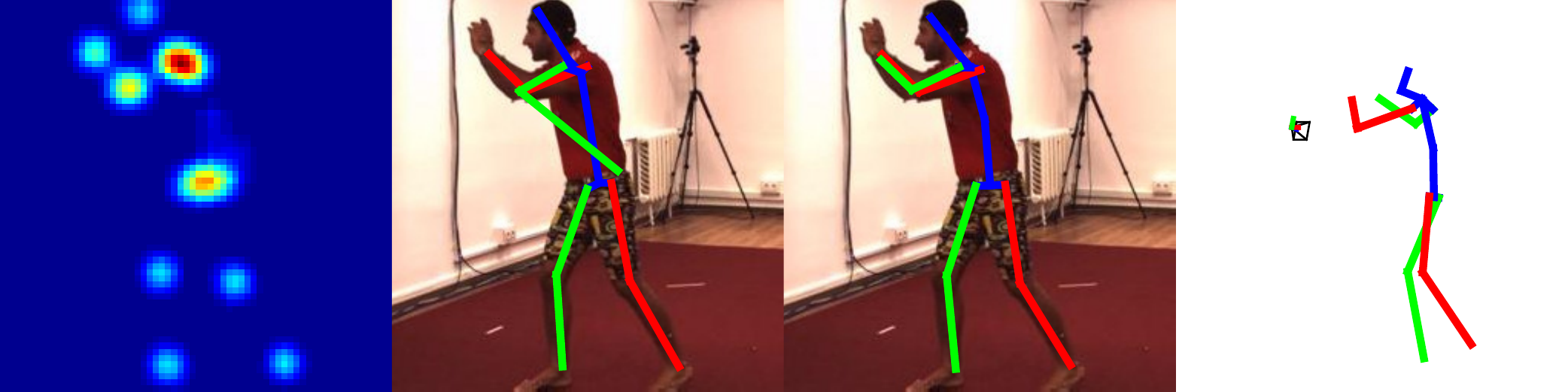}\vspace{0.3em}
  \includegraphics[width=\linewidth]{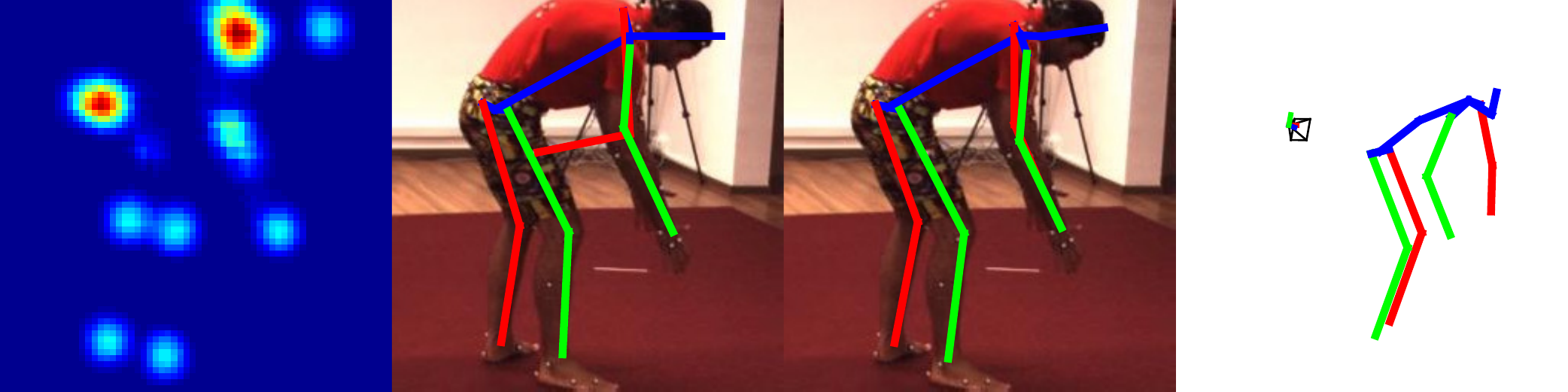}\vspace{0.3em}
  \includegraphics[width=\linewidth]{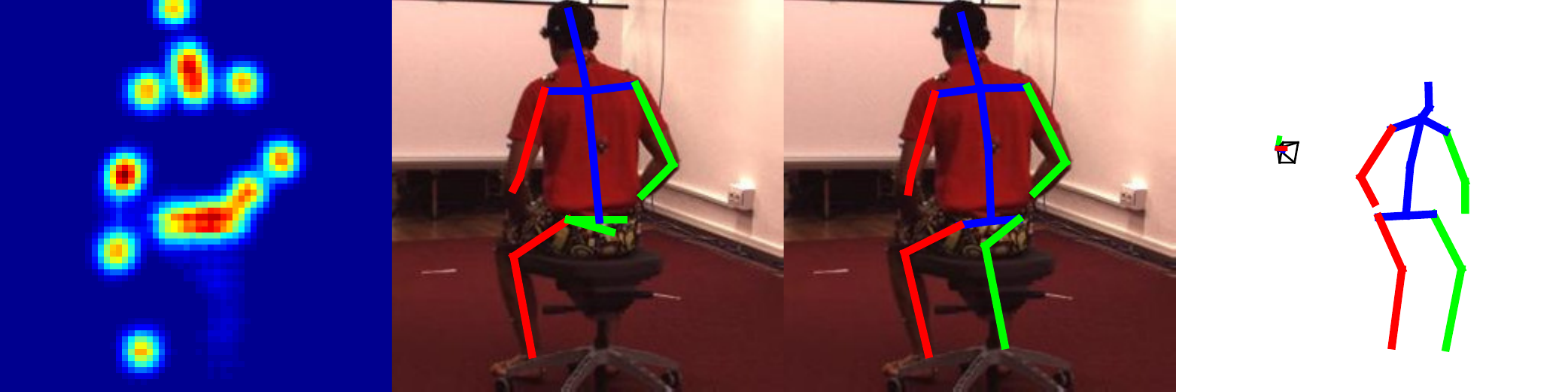}\vspace{0.3em}
  \includegraphics[width=\linewidth]{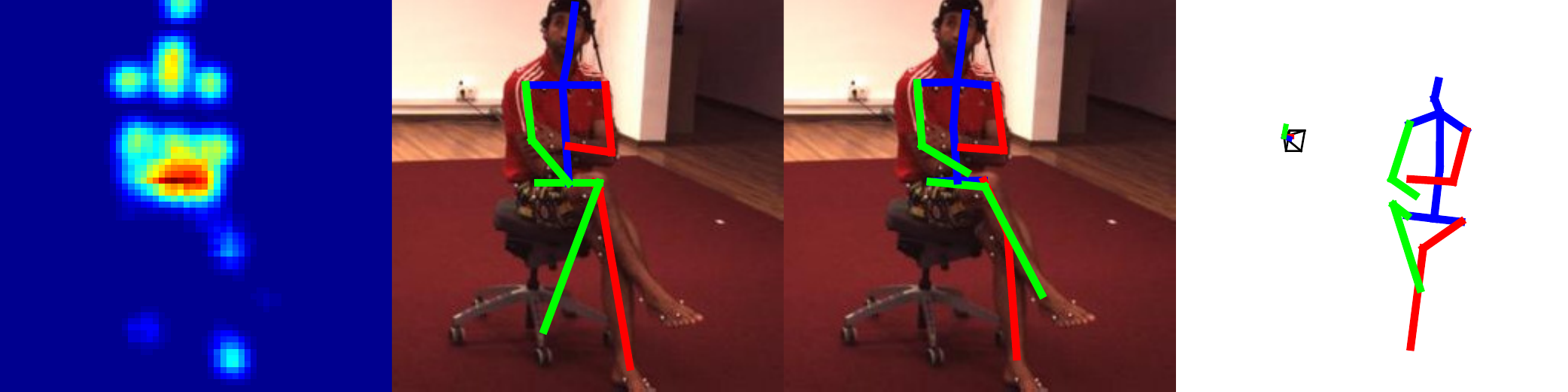}\vspace{0.3em}
  \includegraphics[width=\linewidth]{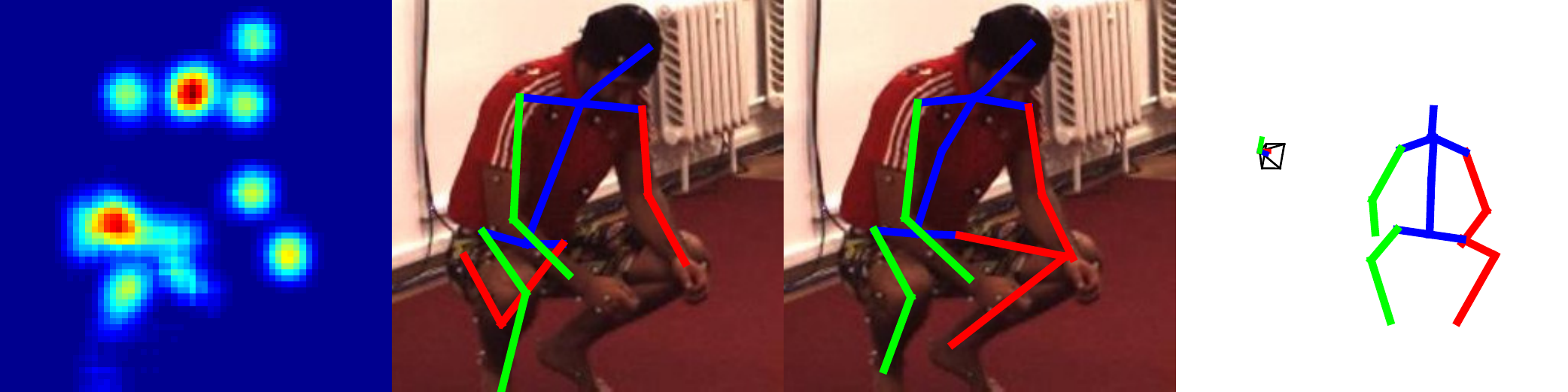}\vspace{0.3em}
  \includegraphics[width=\linewidth]{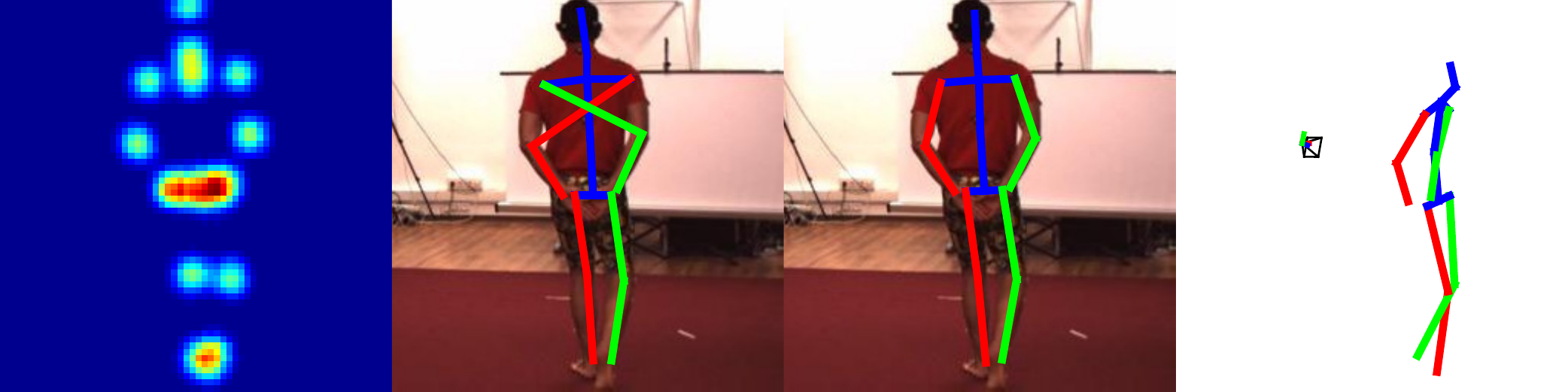}\vspace{0.3em}
  \end{minipage}
  \begin{minipage}{0.49\textwidth}
  \includegraphics[width=\linewidth]{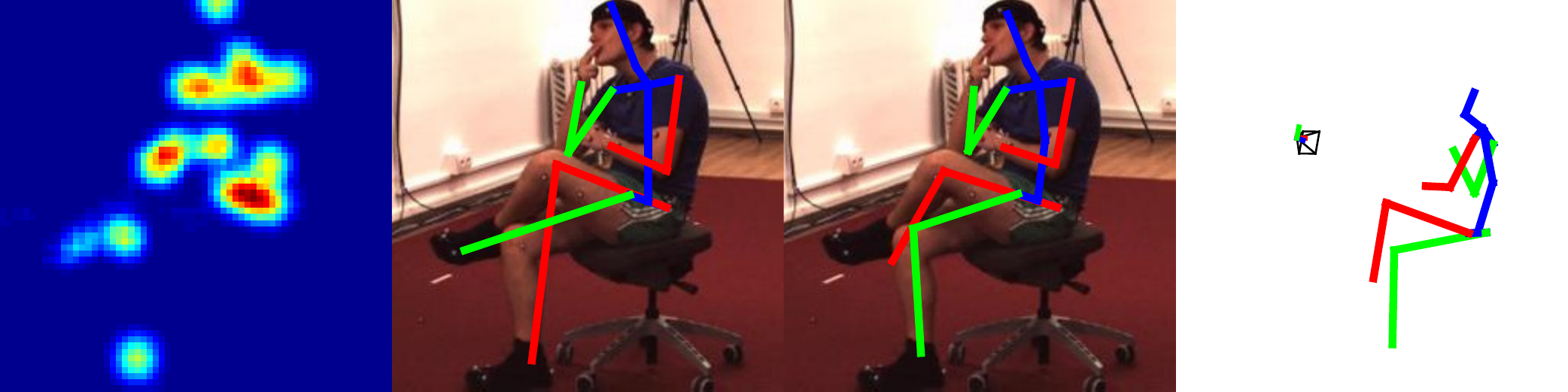}\vspace{0.3em}
  \includegraphics[width=\linewidth]{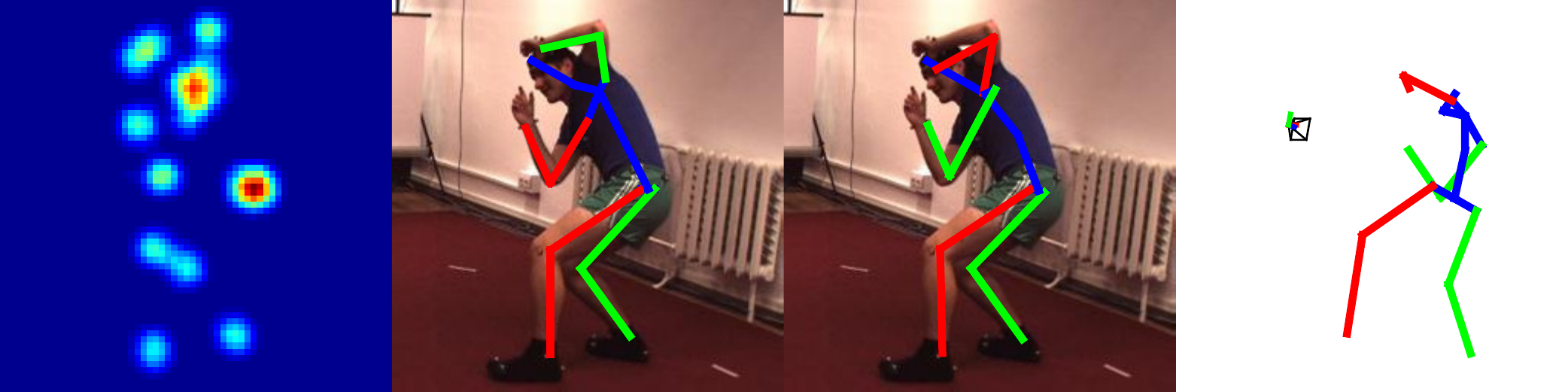}\vspace{0.3em}
  \includegraphics[width=\linewidth]{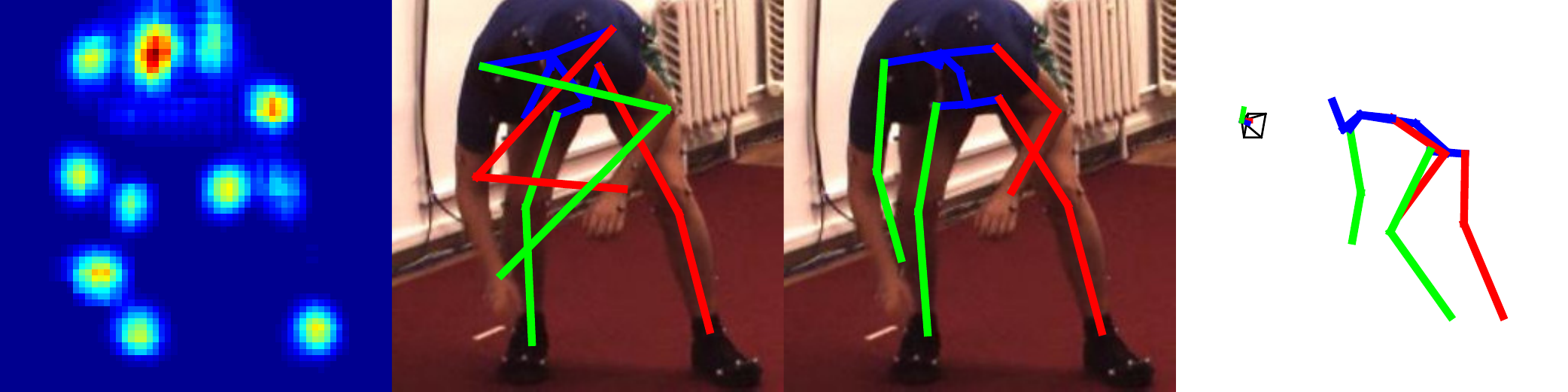}\vspace{0.3em}
  \includegraphics[width=\linewidth]{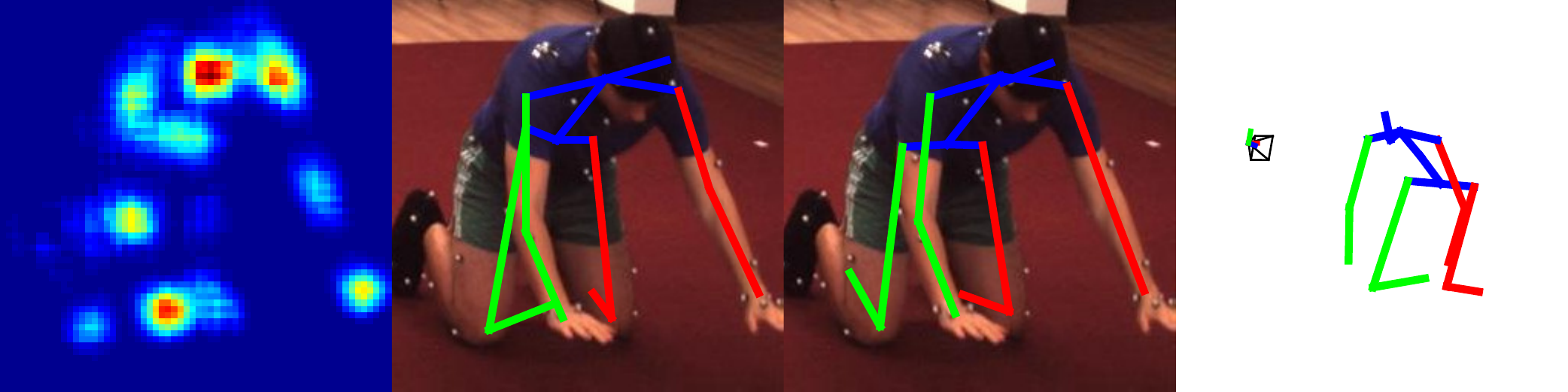}\vspace{0.3em}
  \includegraphics[width=\linewidth]{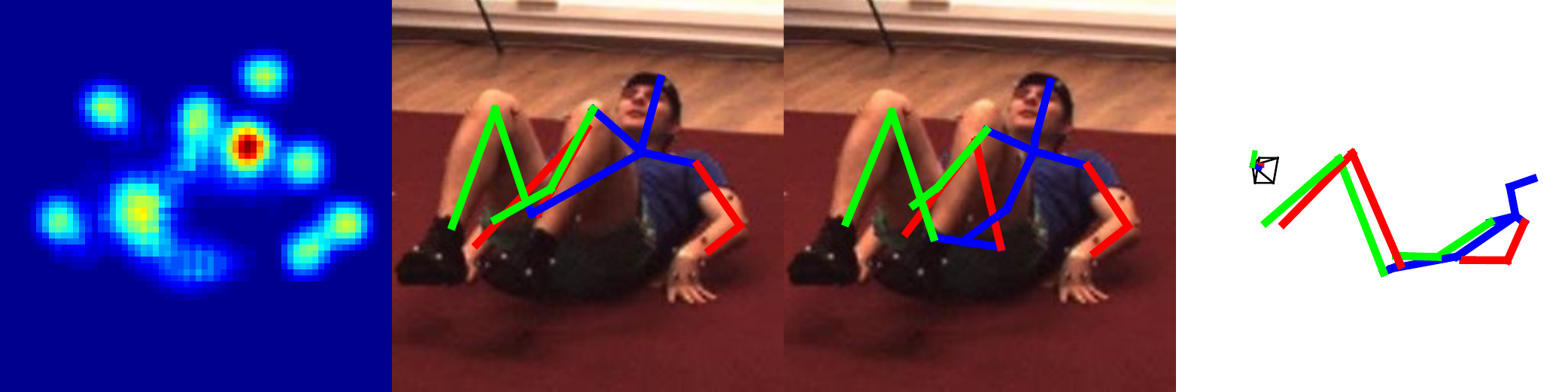}\vspace{0.3em}
  \includegraphics[width=\linewidth]{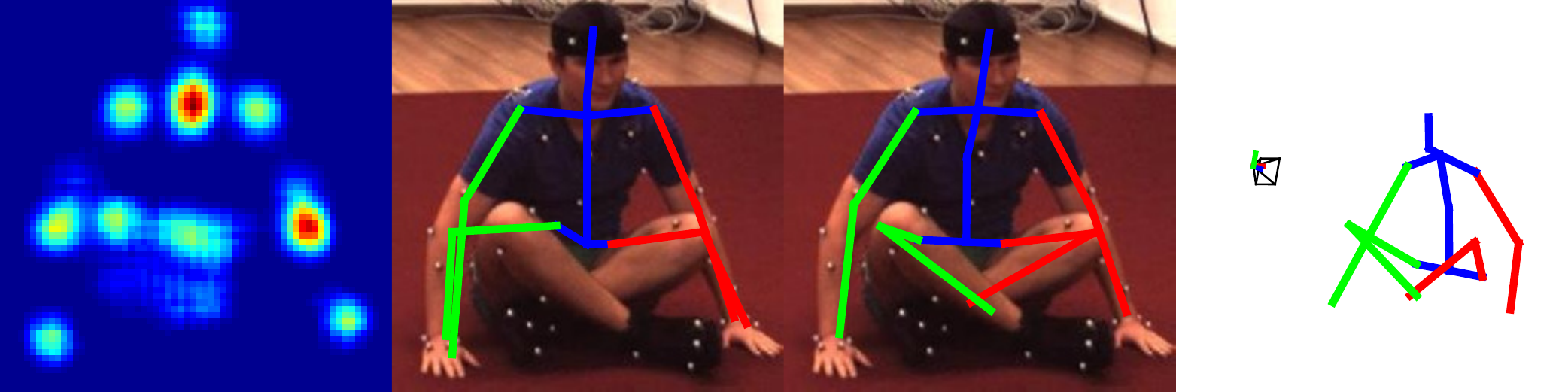}\vspace{0.3em}
  \end{minipage}
  \caption{Example comparative frame results on Human3.6M \cite{ionescu2014human}. Each row includes two examples. The images from left-to-right correspond to the heat map (all joints shown simultaneously), the 2D pose found by greedily locating each joint separately according to the heat map, the estimated 2D pose by the proposed EM algorithm, and the estimated 3D pose visualized in a novel view. The original viewpoint is also shown. Notice that the errors in the 2D heat maps are corrected after considering the pose and temporal smoothness priors. }\label{fig:h36m}
\end{figure*}

\begin{figure*}
  \centering
  \begin{minipage}{0.48\textwidth}
  \includegraphics[width=0.95\linewidth]{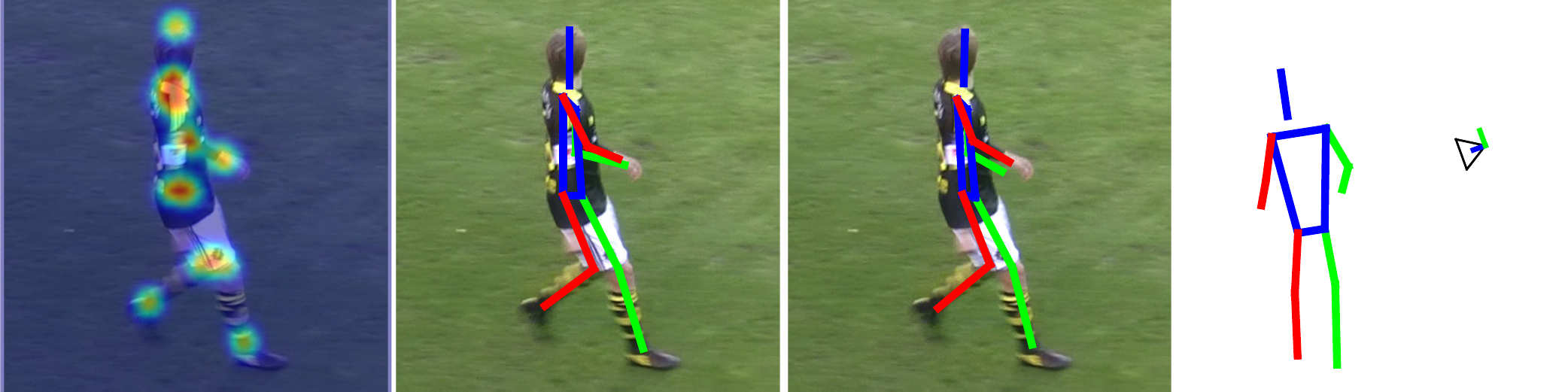}
  \includegraphics[width=0.95\linewidth]{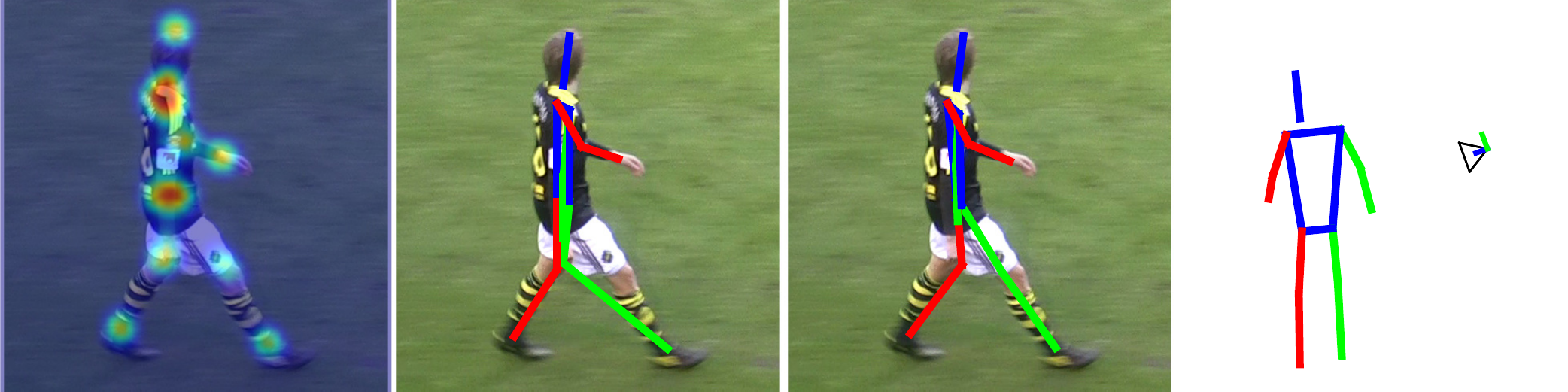}
  \includegraphics[width=0.95\linewidth]{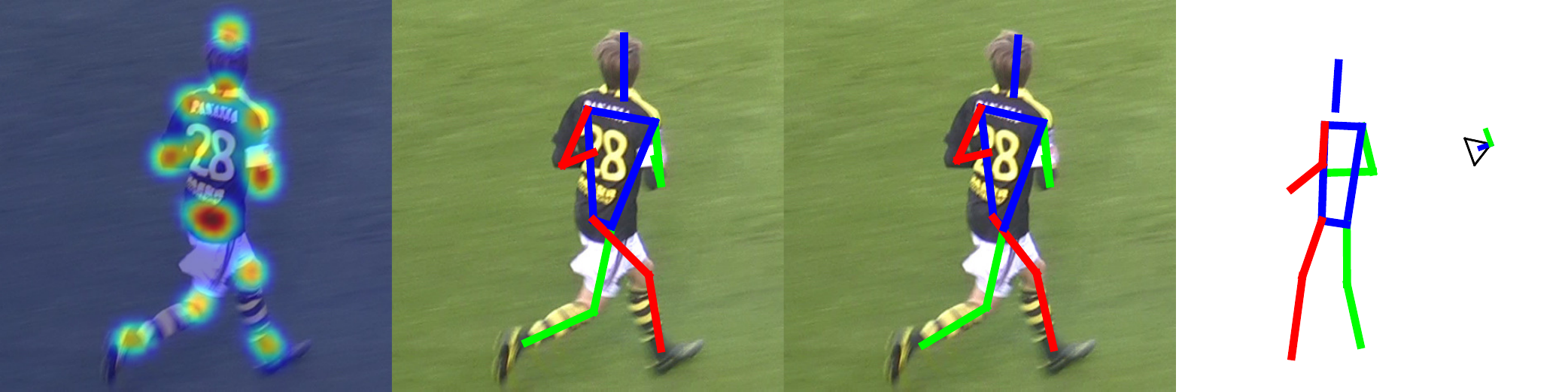}
  \end{minipage}
  \hfill
  \begin{minipage}{0.48\textwidth}
  \includegraphics[width=0.95\linewidth]{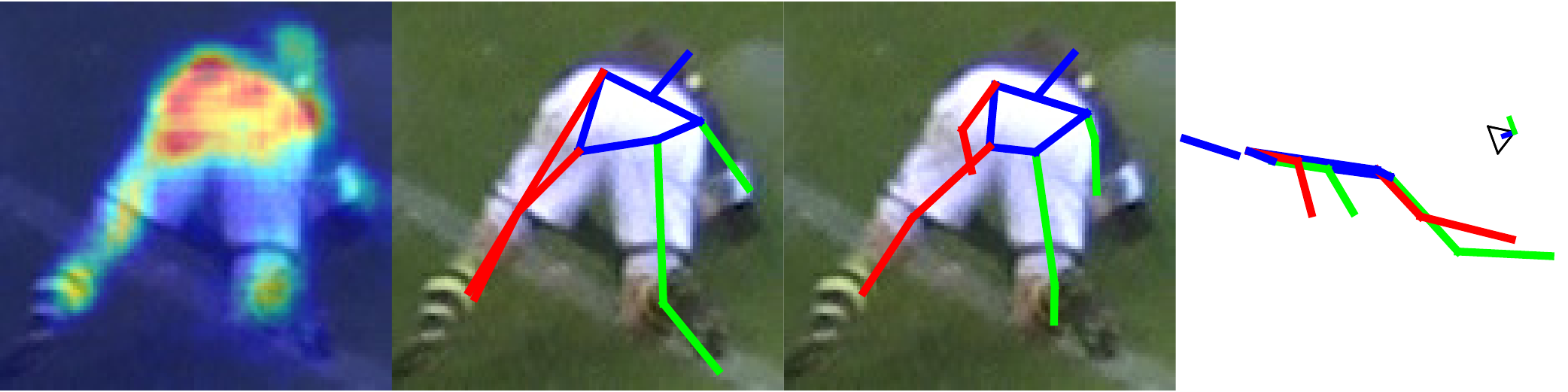}
  \includegraphics[width=0.95\linewidth]{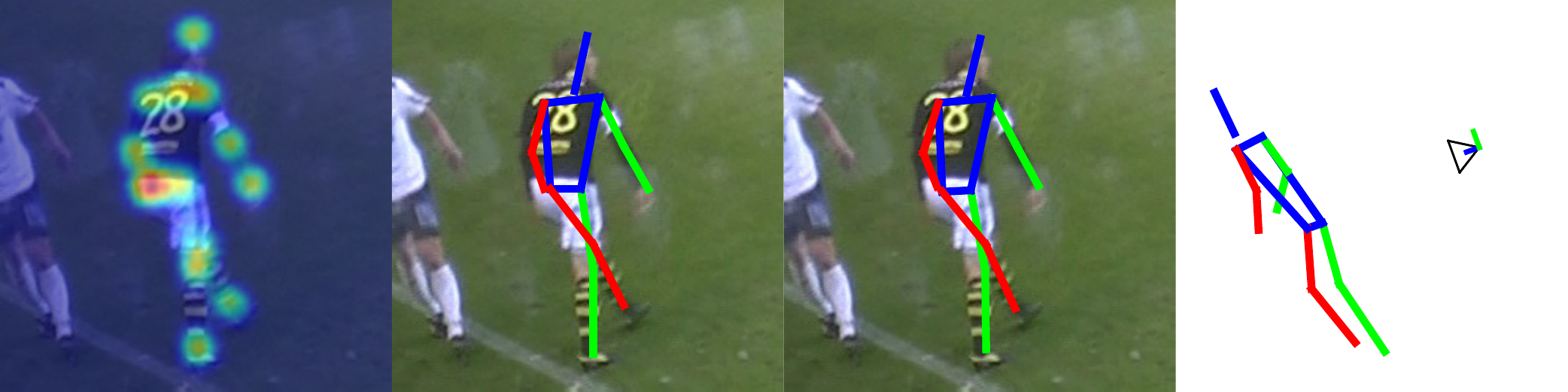}
  \includegraphics[width=0.95\linewidth]{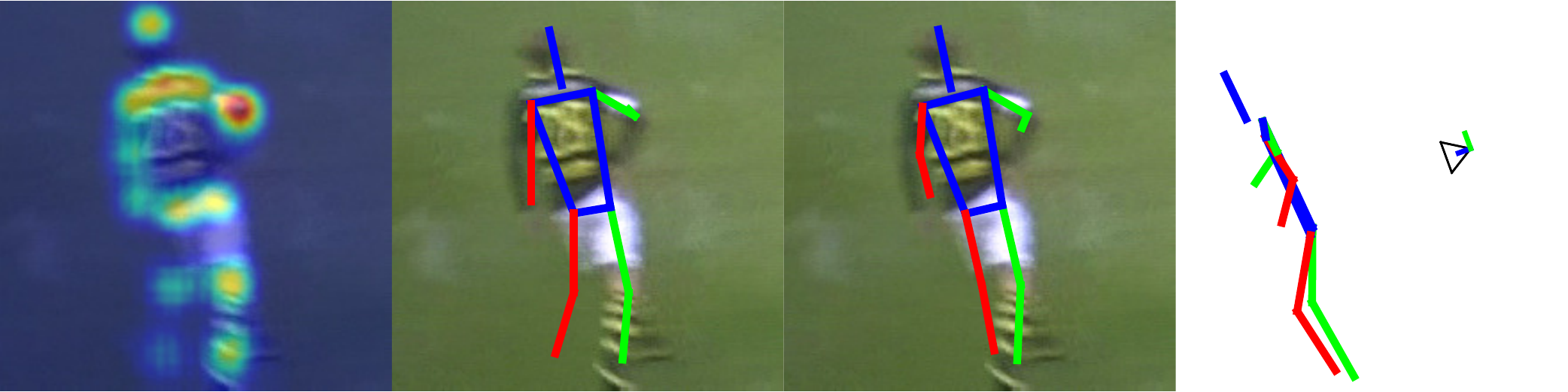}
  \end{minipage}
  \caption{Example comparative frame results on KTH Football II \cite{burenius20133d}.
 The images from left-to-right in each example correspond to the heat map (all joints shown simultaneously), the 2D pose found by greedily locating each joint separately according to the heat map response, the estimated 2D pose by the proposed EM algorithm, and the estimated 3D pose visualized in a novel view. The original viewpoint is also shown. }\label{fig:kth}
\end{figure*}

\begin{table*}
\caption{Mean reconstruction errors (mm) on HumanEva I \cite{sigal2010humaneva}.}
\centering
\renewcommand{\arraystretch}{1.5}
\begin{tabular}{l*{3}{c}*{4}{c}}
\toprule
& \multicolumn{3}{c}{Walking} & \multicolumn{3}{c}{Jogging} & Average \\
& S1 & S2 & S3 & S1 & S2 & S3 \\
\toprule
Radwan et al. \cite{radwan2013monocular} & 75.1 & 99.8 & 93.8 & 79.2 & 89.8 & 99.4 & 89.5 \\
Wang et al. \cite{wang2014robust} & 71.9 & 75.7 & 85.3 & 62.6 & 77.7 & 54.4 & 71.3 \\
Simo-Serra et al. \cite{simo2013joint} & 65.1 & 48.6 & 73.5 & 74.2 & 46.6 & 32.2 & 56.7\\
Bo et al. \cite{bo2010twin} & 46.4 & {30.3} & 64.9 & 64.5 & 48.0 & 38.2 & 48.7 \\
Kostrikov et al. \cite{kostrikov2014depth} & 44.0 & 30.9 & 41.7 & 57.2 & 35.0 & 33.3 & 40.3 \\
Yasin et al. \cite{yasin2016dual} & 35.8 & 32.4 & {41.6} & {46.6} & 41.4 & 35.4 & 38.9 \\
Proposed & {34.3} & 31.6 & 49.3 & 48.6 & {34.0} & {30.0} & {37.9}\\
\toprule
\end{tabular}
\label{tab:humaneva}
\end{table*}

\refFig{fig:par} shows the mean reconstruction error as a function of each parameter in \refEq{eq:prior} while fixing the others, evaluated on a subset of test sequences (first five actions from S9). The first curve shows that the error initially decreases rapidly when $\alpha$ becomes nonzero.  This indicates the importance of the sparsity constraint. After the initial rapid decrease, the error changes very smoothly, which indicates that the solution is not very sensitive to $\alpha$ when its value is in a proper range. Similar observations are made for the weights of the smoothness terms, $\beta$ and $\gamma$. 
In practice, the model parameters were fixed without specific tuning in all other experiments presented in this paper ($\alpha=0.5$, $\beta=20$ and $\gamma=2$ in a normalized 2D coordinate system).

\subsubsection{Qualitative illustration}

\refFig{fig:h36m} visualizes the results on several example frames. While the heat maps may be erroneous due to occlusion, left-right ambiguity, and other sources of uncertainty from the detectors, the proposed EM algorithm can effectively correct the errors by leveraging the pose prior, integrating temporal smoothness, and modeling the uncertainty.

\subsection{HumanEva I}

In this section, the evaluation results on HumanEva I \cite{sigal2010humaneva} are presented. The evaluation protocol described elsewhere \cite{simo2013joint} was adopted. The walking and jogging sequences from camera C1 of all subjects were used for evaluation. The 2D joint detector trained on Human3.6M was fine-tuned with the training sequences for each action separately. Action specific pose dictionaries were learned for each subject separately. Each 3D pose reconstructed by the proposed approach was scaled to have the same average limb length as the training data.

The mean reconstruction errors for the evaluation sequences are reported in \refTab{tab:humaneva}. The results of the compared baselines are taken from prior work \cite{yasin2016dual}. Due to the large overlap between training and test data and less variability of poses, 
higher accuracies 
are generally obtained on this dataset compared to Human3.6M for all approaches. While none of the approaches dominate across all sequences, ours achieves the best overall accuracy.

\subsection{KTH Football II}

KTH Multiview Football II \cite{kazemi2013multi} contains images of professional footballers playing a match.
It includes image sequences with 3D ground truth for 14 annotated joints captured from three calibrated views.
The 3D ground truth was generated using multiview reconstruction with manual 2D annotations.
Our evaluation was performed using the standard protocol \cite{tekin2015predicting}, where the sequences of ``Player 2'' from ``Camera 1" were used for testing. The generic hourglass model trained on MPII was used as the 2D detector without fine-tuning, while the pose dictionary was learned using the 3D poses associated with the training images provided in this dataset.
Each 3D pose reconstructed by the proposed approach was scaled to have the same average limb length as the training poses and then aligned to the ground truth by a translation according to the root location.

To compare with the baselines, reported results are based on the percentage of correct pose (PCP) to measure part localization in 3D.
\refTab{tab:kth} presents a summary of PCP results.  It shows that the proposed approach achieves improved accuracy over the state-of-the-art. Results on selected frames are visualized in \refFig{fig:kth}.

\begin{table}
\caption{Mean PCP scores on KTH Football II \cite{burenius20133d}.}
\centering
\renewcommand{\arraystretch}{1.5}
\begin{tabular}{l*{3}{c}*{1}{c}}
\toprule
& \multicolumn{3}{c}{Sequence 1} & \multicolumn{1}{c}{Sequence 2} \\
& \cite{burenius20133d} & \cite{tekin2015predicting} & Proposed & Proposed \\
\toprule
Upper Arms & 14 & 74 & {89} & 61\\
Lower Arms & 06 & 49 & {78} & 49\\
Upper Legs & 63 & 98 & {99} & 77\\
Lower Legs & 41 & 77 & {85} & 56\\
\toprule
\end{tabular}

\label{tab:kth}
\end{table}

\subsection{MPII}

\begin{figure*}
  \centering
  \includegraphics[width=0.48\linewidth]{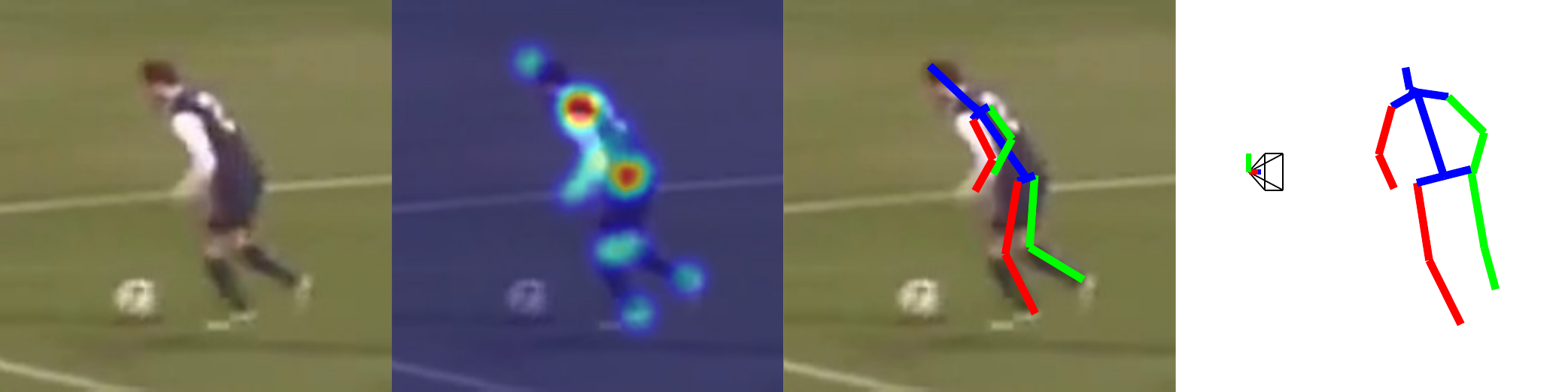}
  \includegraphics[width=0.48\linewidth]{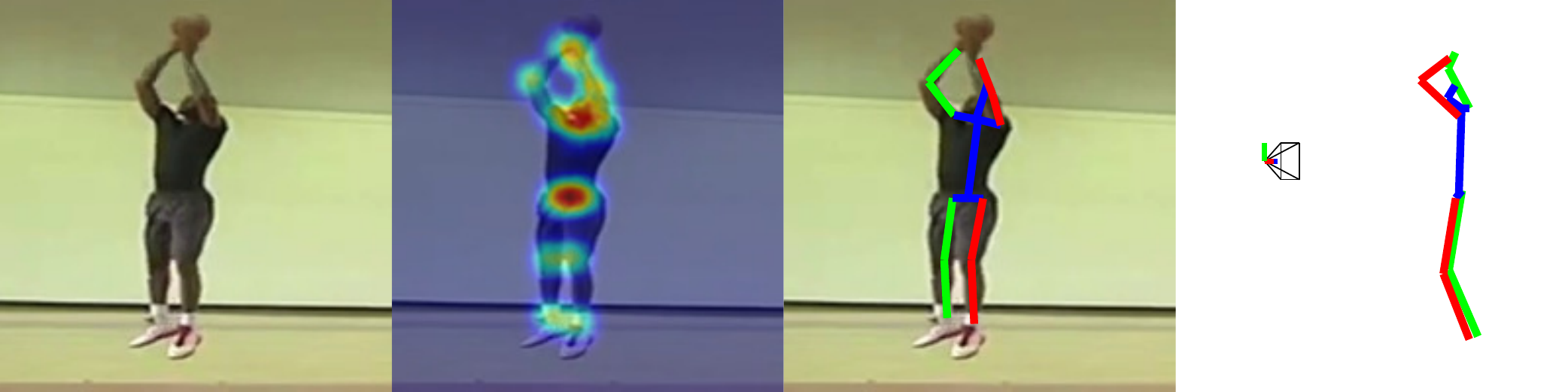}
  \includegraphics[width=0.48\linewidth]{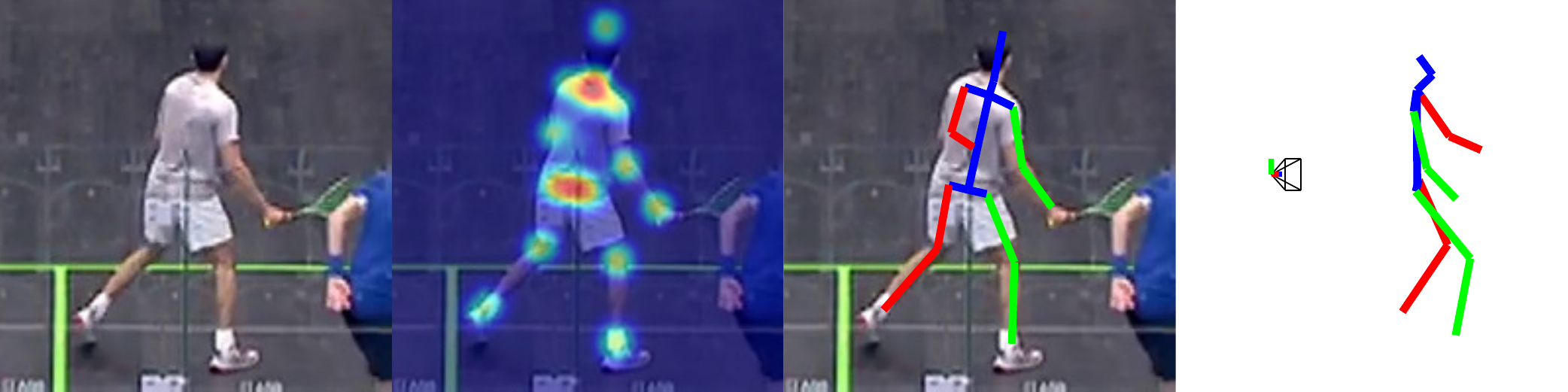}
  \includegraphics[width=0.48\linewidth]{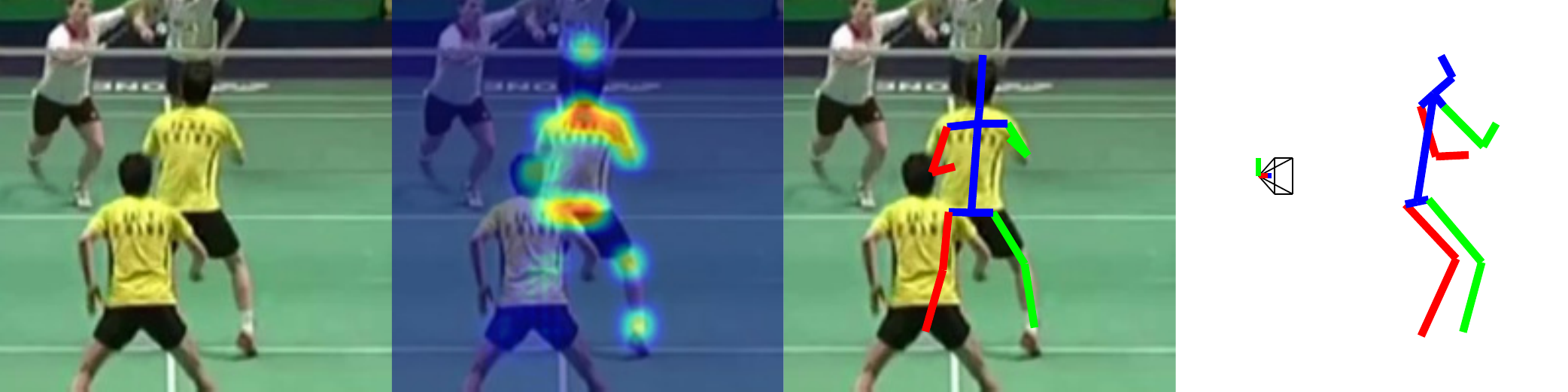}
  \includegraphics[width=0.48\linewidth]{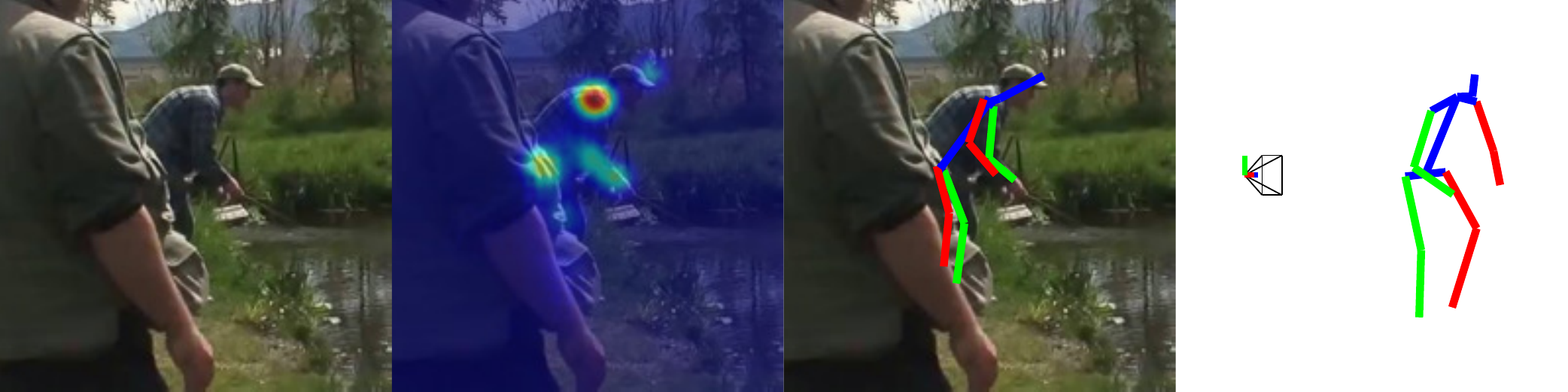}
  \includegraphics[width=0.48\linewidth]{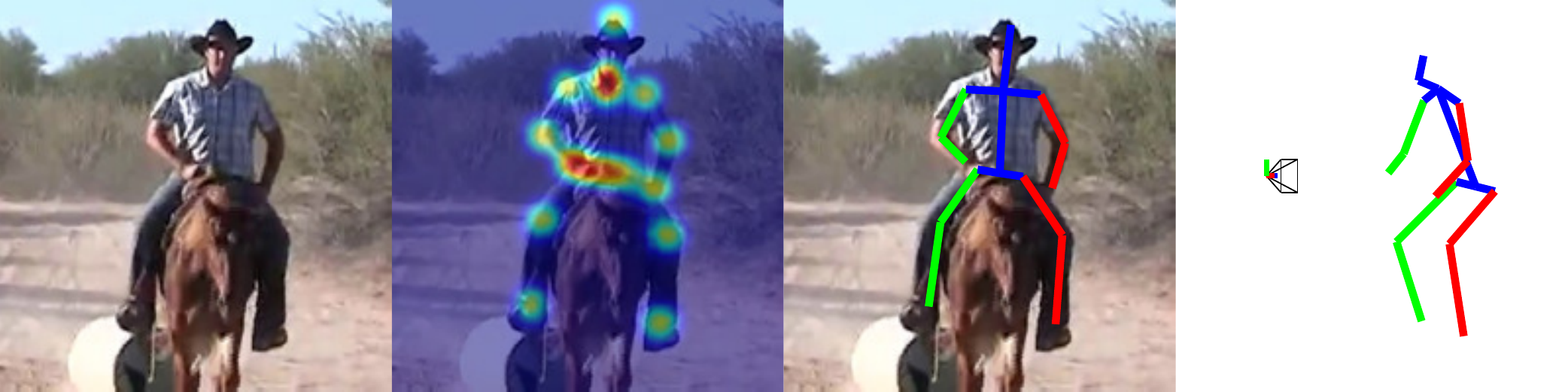}
  \includegraphics[width=0.48\linewidth]{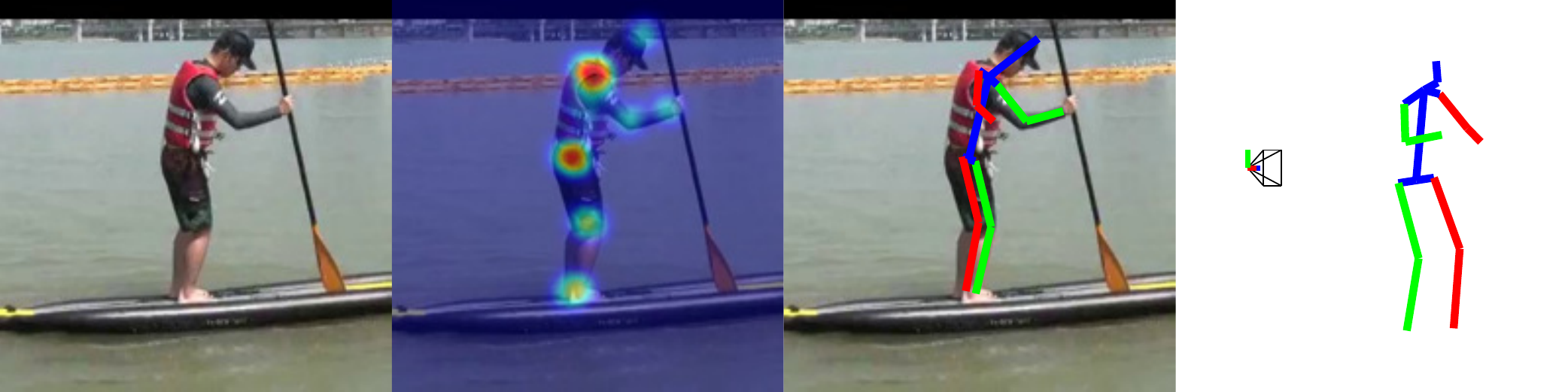}
  \includegraphics[width=0.48\linewidth]{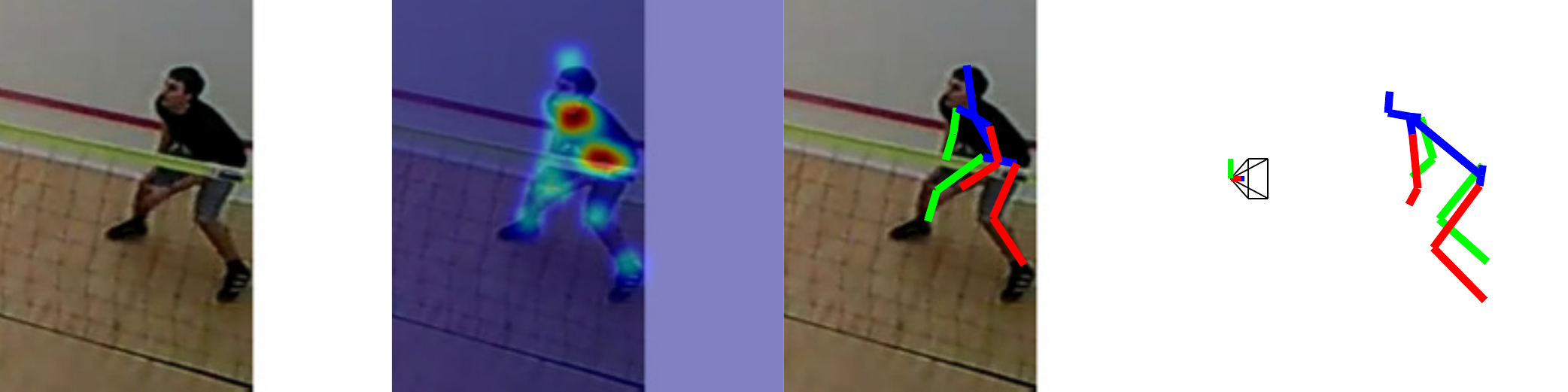}
  \includegraphics[width=0.48\linewidth]{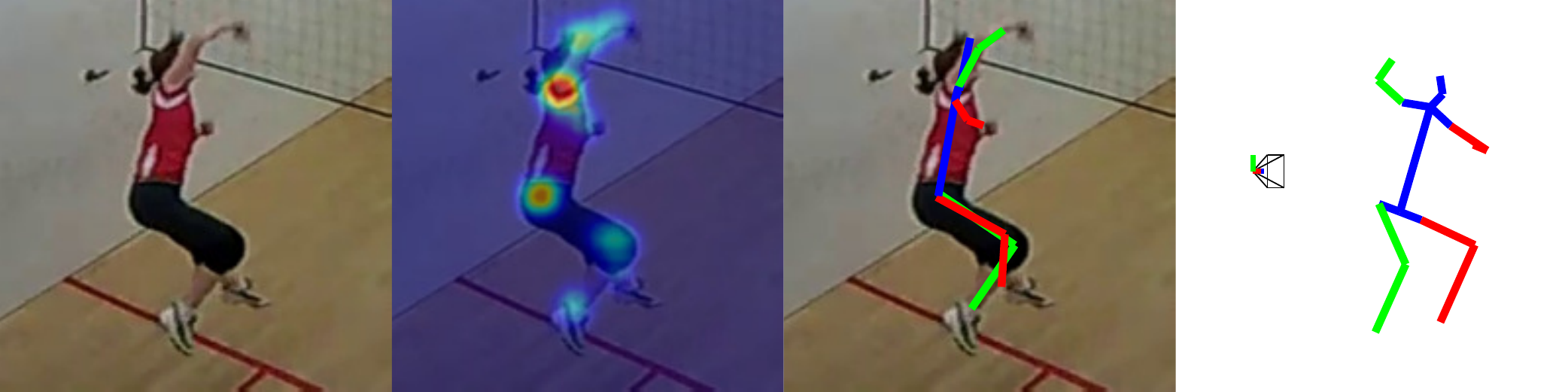}
  \includegraphics[width=0.48\linewidth]{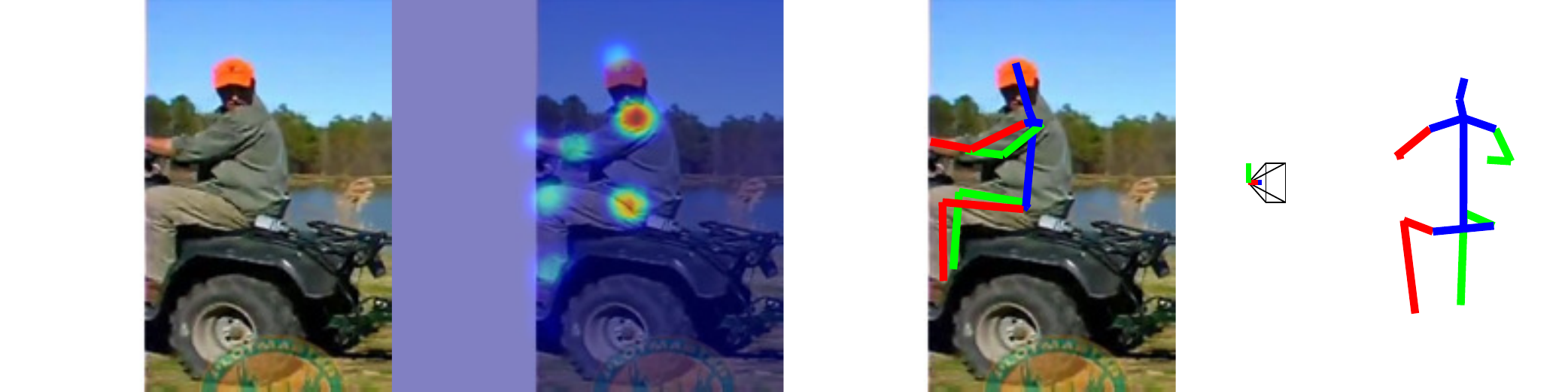}
  \includegraphics[width=0.48\linewidth]{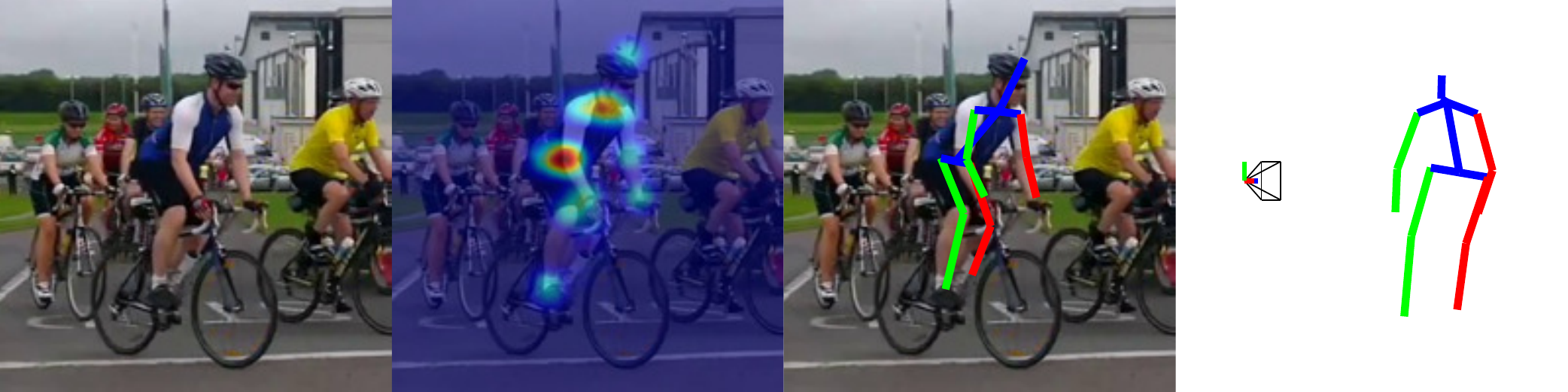}
  \includegraphics[width=0.48\linewidth]{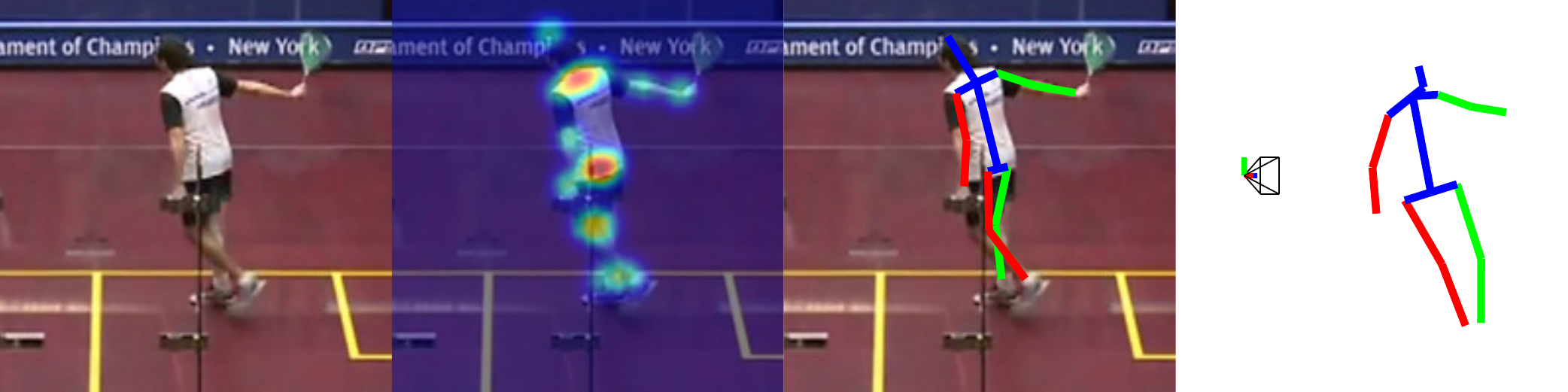}
  \includegraphics[width=0.48\linewidth]{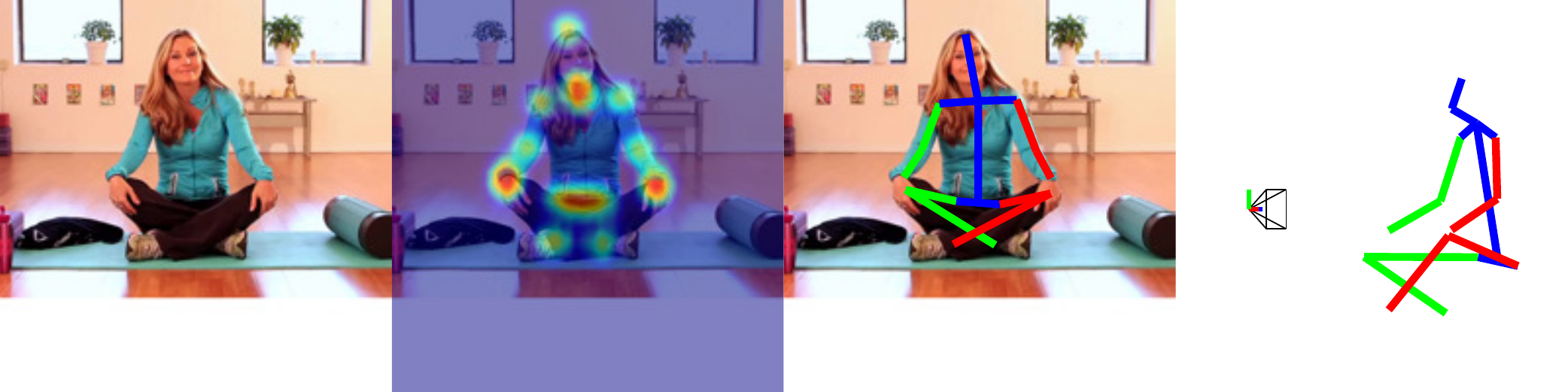}
  \includegraphics[width=0.48\linewidth]{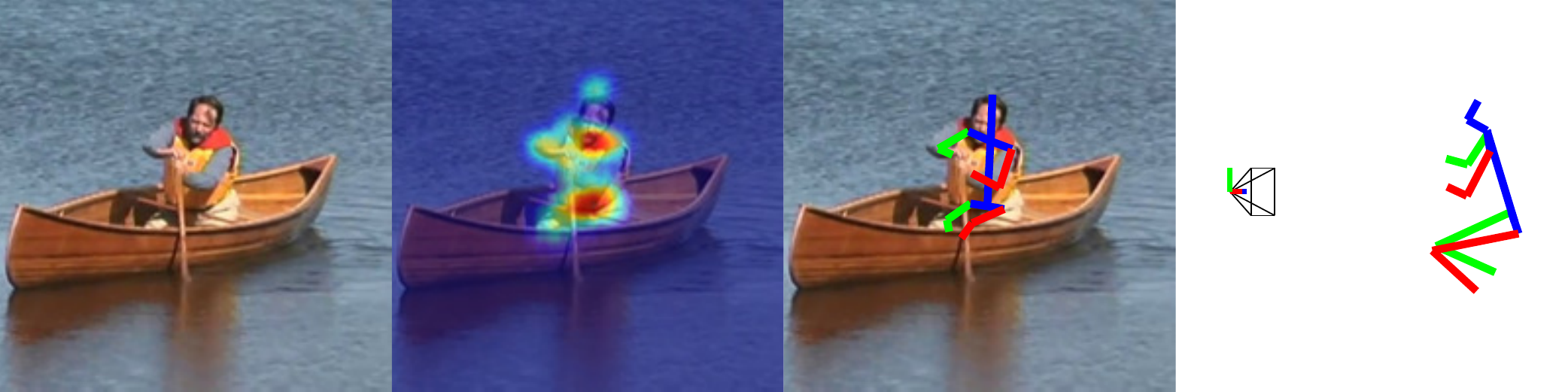}
  \caption{Example successes on MPII \cite{andriluka14cvpr}. In each example, the images from left-to-right correspond to the input image, the heat map (all joints shown simultaneously), the estimated 2D pose, and the estimated 3D pose visualized in a novel view. The original viewpoint is also shown.}\label{fig:mpii-good}
\end{figure*}

Finally, the applicability of our proposed approach to in-the-wild imagery
is qualitatively illustrated with the MPII Human Pose dataset \cite{andriluka14cvpr}.  MPII is a large-scale 2D human pose dataset that includes 25K single images extracted from YouTube videos containing over 40K people and 410 activities.  This dataset does not include 3D pose data. 
The original hourglass model \cite{newell2016stacked} trained on this dataset was used as the 2D detector and combined with the nonspecific action pose dictionary learned on Human3.6M to reconstruct the 3D human poses. The test images are from the validation set defined in previous work \cite{newell2016stacked}.

\begin{figure*}
  \centering
  \includegraphics[width=0.48\linewidth]{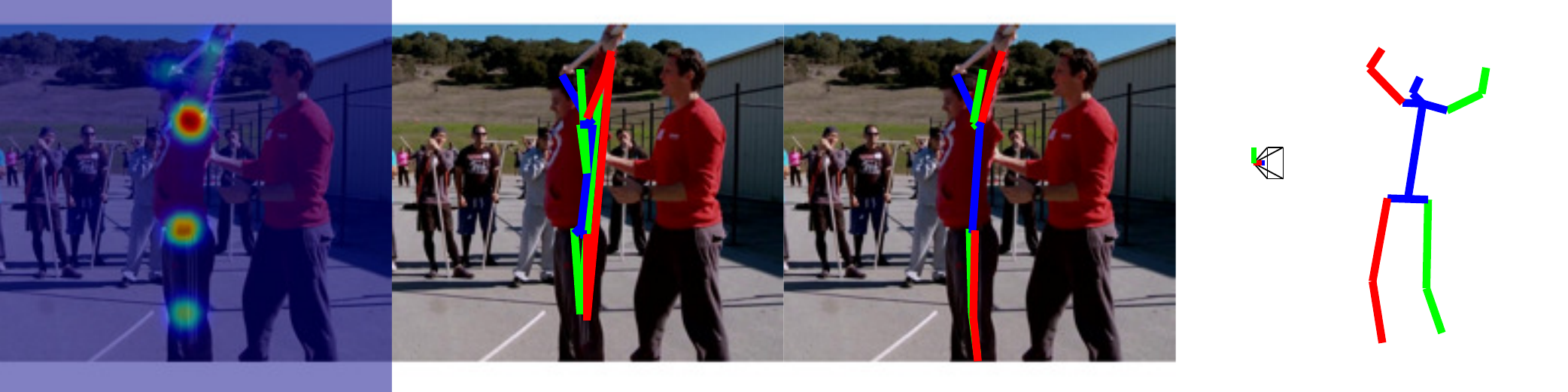}
  \includegraphics[width=0.48\linewidth]{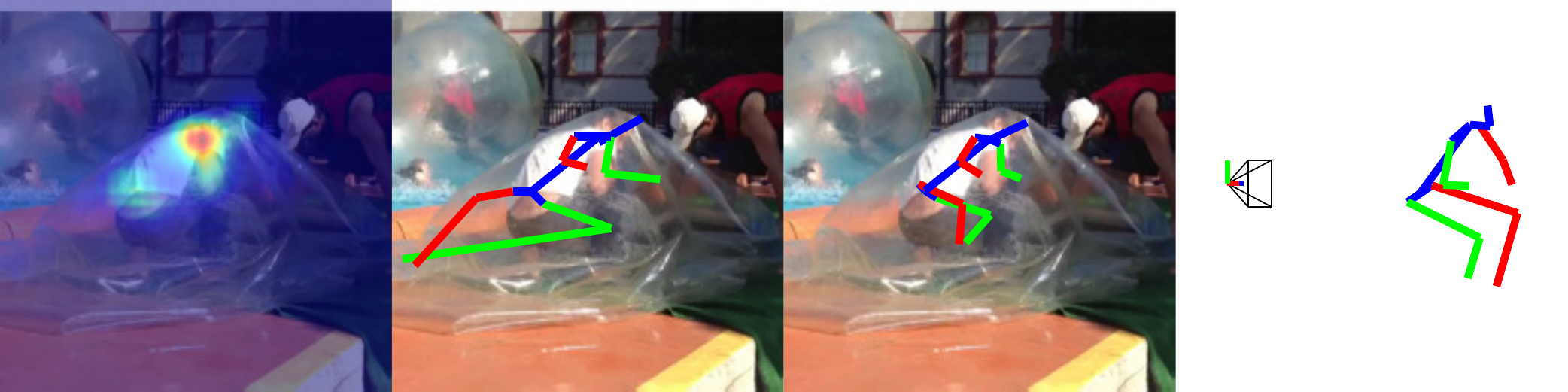}
  \includegraphics[width=0.48\linewidth]{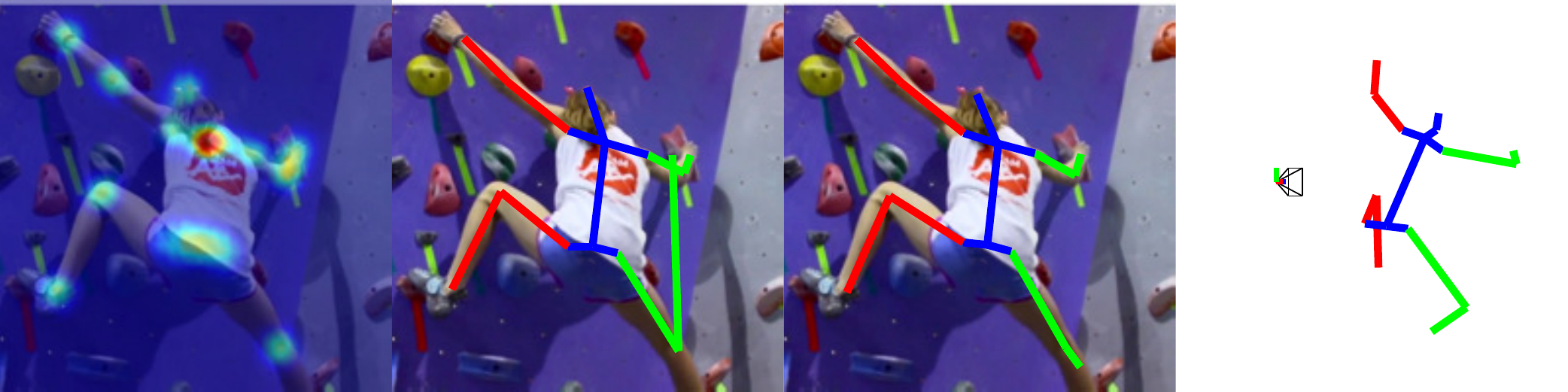}
  \includegraphics[width=0.48\linewidth]{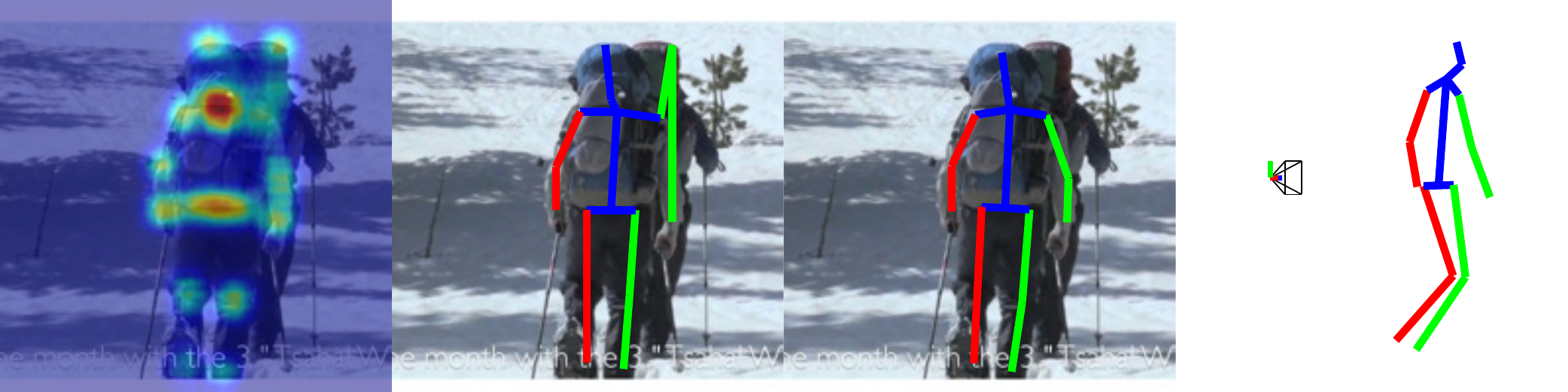}
  \includegraphics[width=0.48\linewidth]{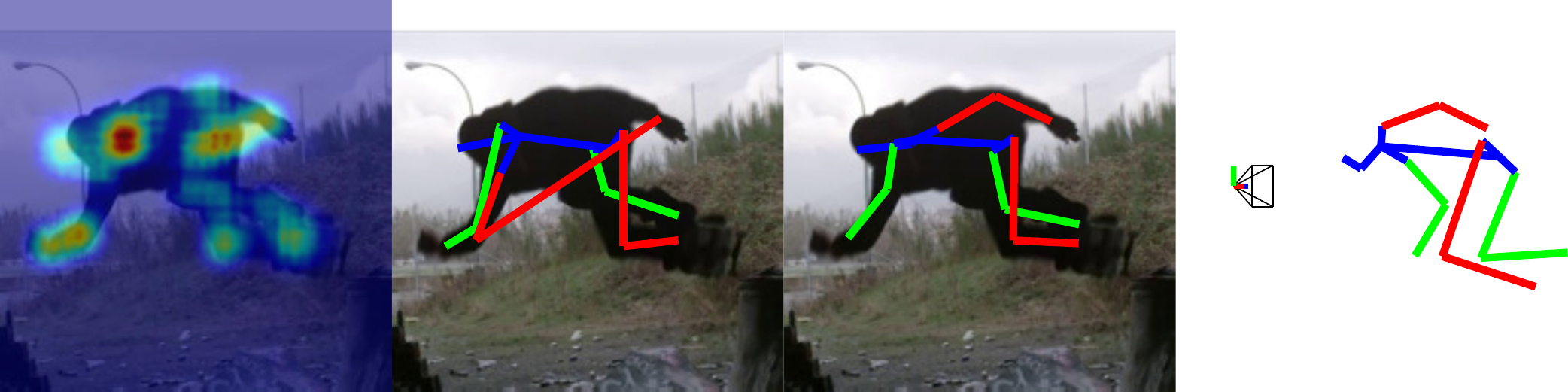}
  \includegraphics[width=0.48\linewidth]{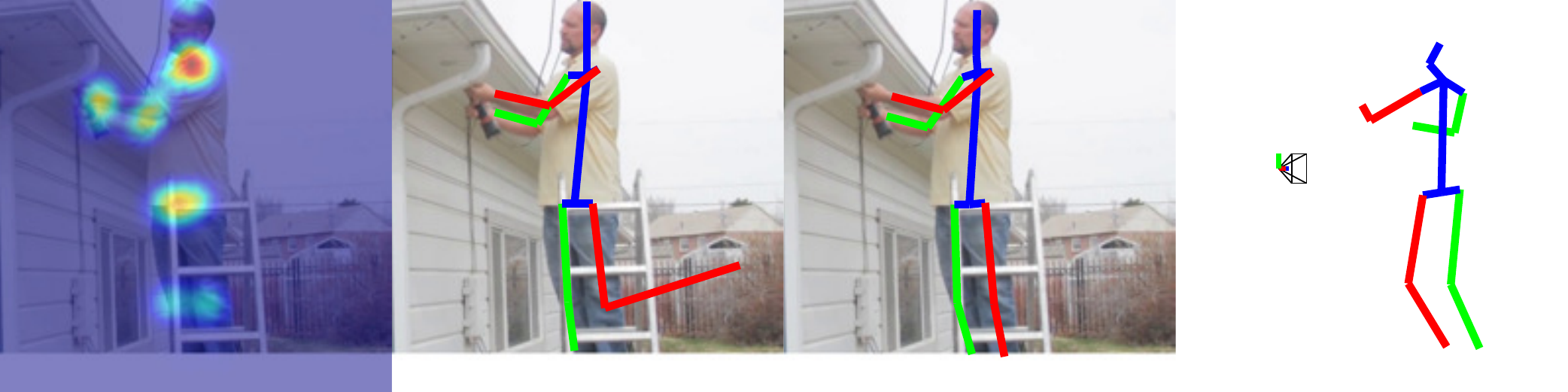}
  \includegraphics[width=0.48\linewidth]{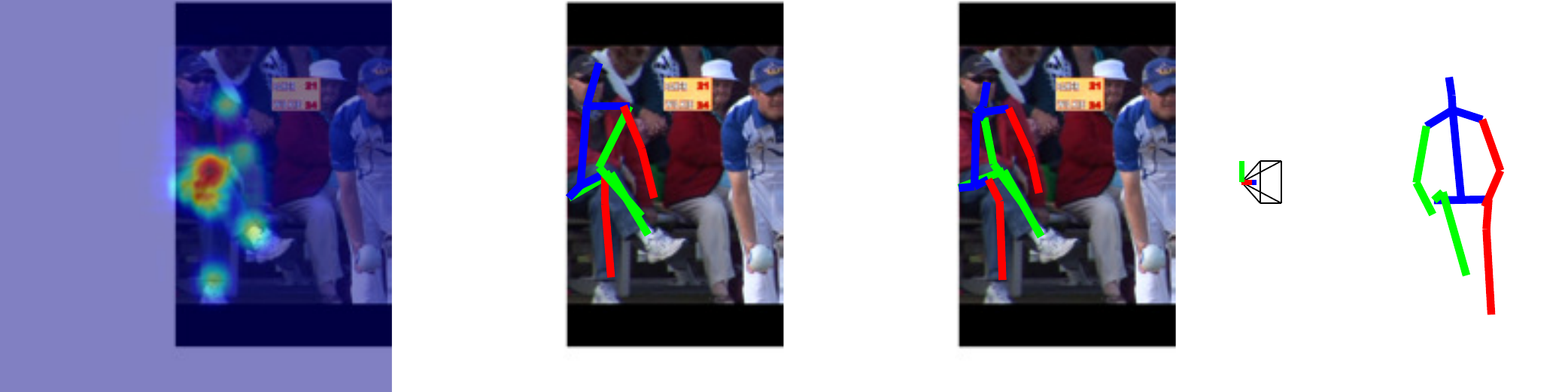}
  \includegraphics[width=0.48\linewidth]{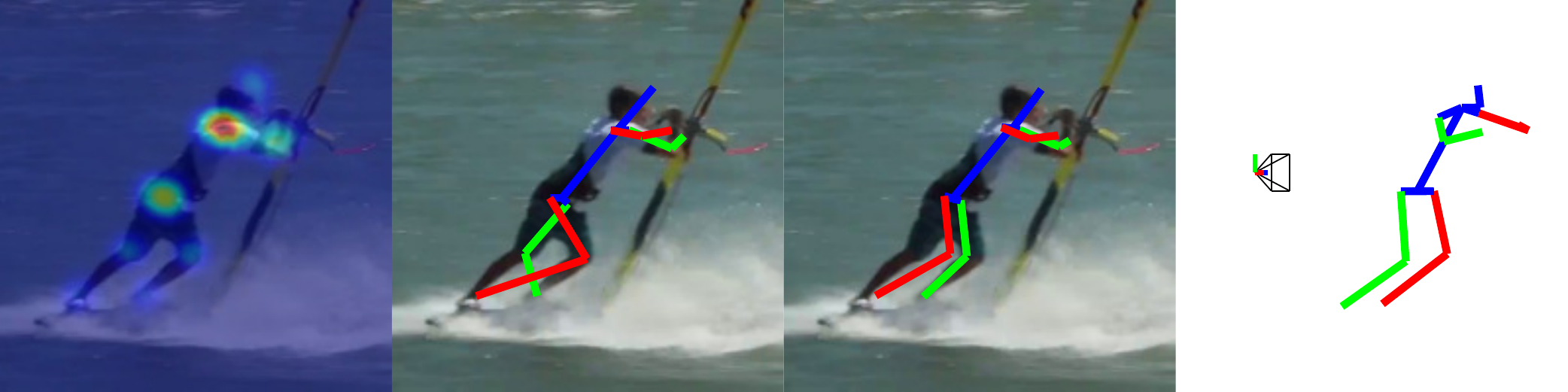}
  \caption{Example comparative frame results on MPII \cite{andriluka14cvpr}.  In each example, the images from left-to-right correspond to the heat map (all joints shown simultaneously), the 2D pose found by greedily locating each joint separately according to the heat map, the estimated 2D pose by the proposed EM algorithm, and the estimated 3D pose visualized in a novel view. The original viewpoint is also shown.  Notice that the errors in the 2D heat maps are corrected after considering the 3D pose prior.}\label{fig:mpii-improve}
\end{figure*}

\begin{figure*}
  \centering
  \includegraphics[width=0.48\linewidth]{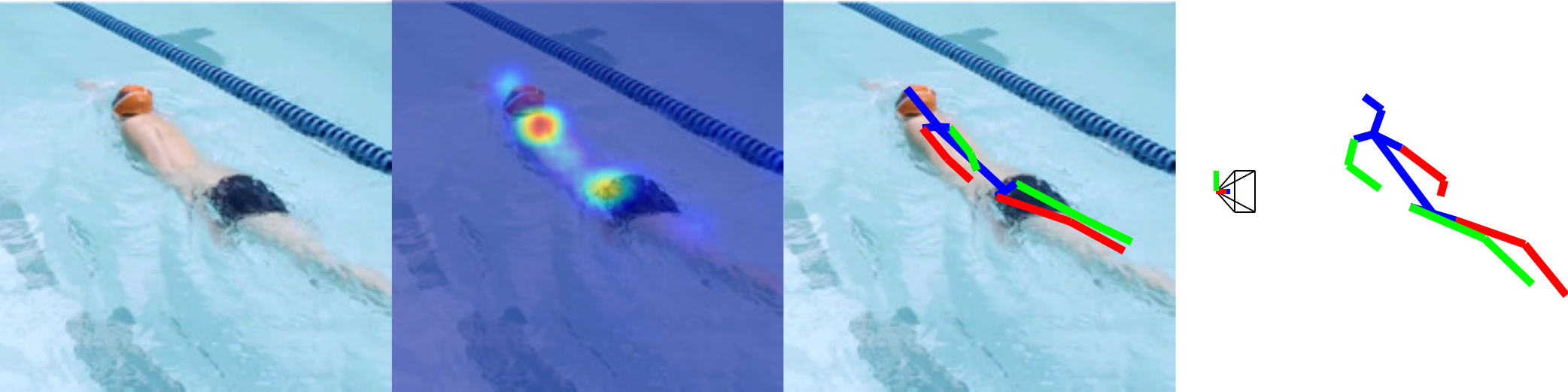}
  \includegraphics[width=0.48\linewidth]{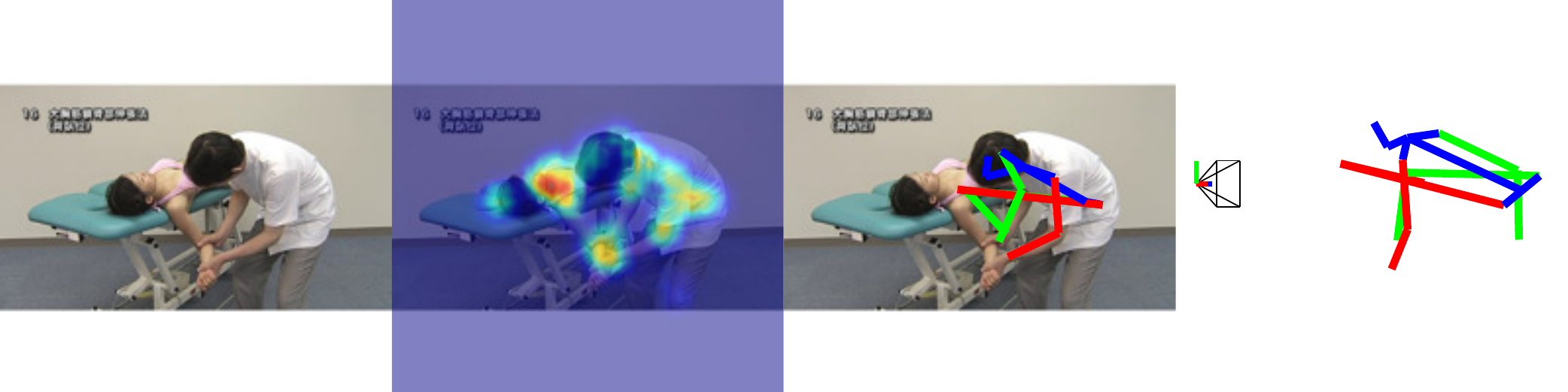}
  \includegraphics[width=0.48\linewidth]{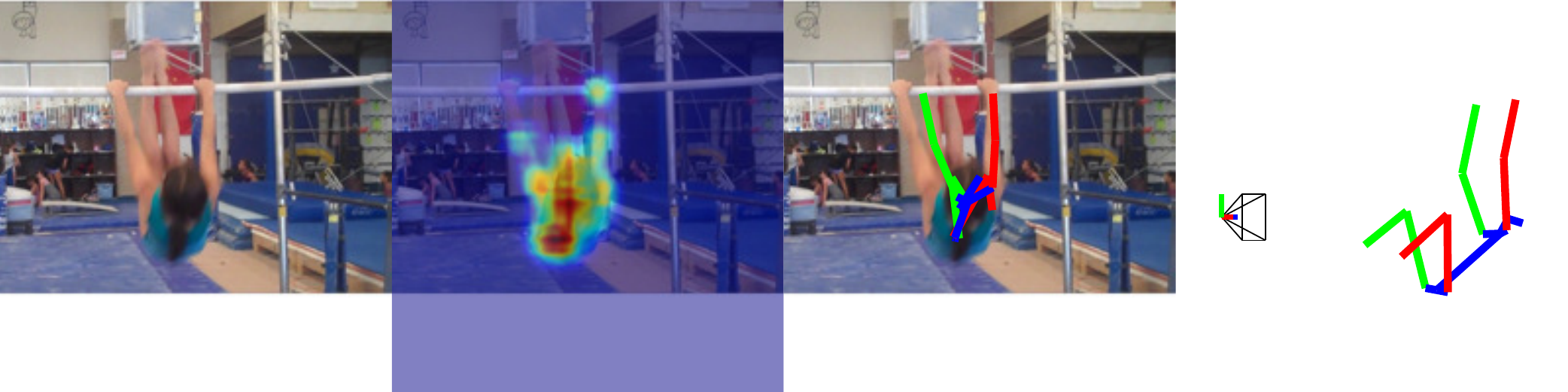}
  \includegraphics[width=0.48\linewidth]{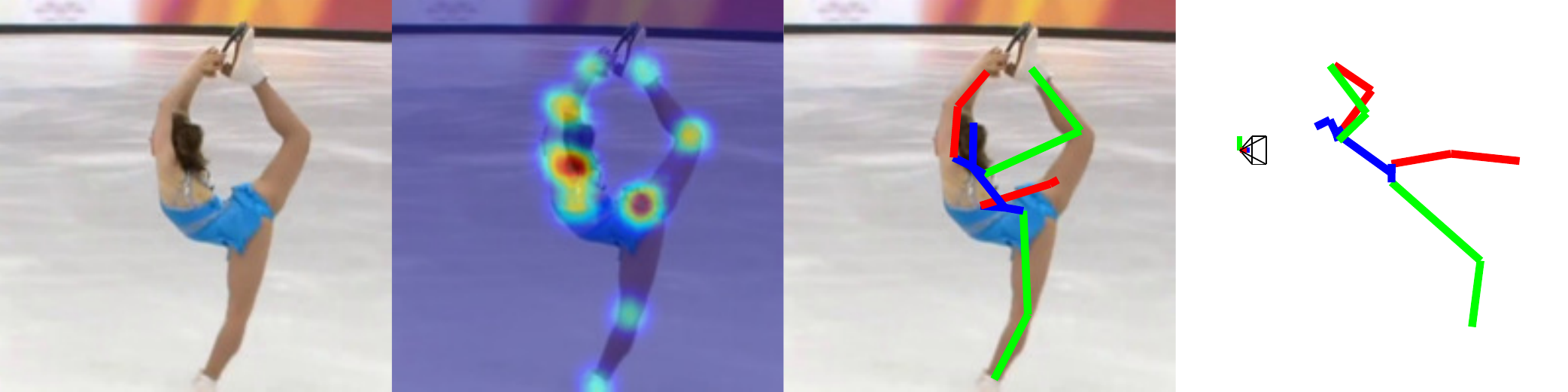}
  \caption{Example failures on MPII \cite{andriluka14cvpr}. In each example, the images from left-to-right correspond to the input image, the heat map (all joints shown simultaneously), the estimated 2D pose, and the estimated 3D pose visualized in a novel view. The original viewpoint is also shown. }\label{fig:mpii-bad}
\end{figure*}

\refFig{fig:mpii-good} shows successful examples on MPII. Note that the input data consists of single images rather than sequences. While the pose dictionary is learned from another dataset, the proposed approach is able to produce visually reasonable 3D reconstructions from single images for a large variety of activities and viewpoints. \refFig{fig:mpii-improve} presents some examples with larger uncertainties in the 2D heat maps, which result in incorrect 2D poses if the joints are simply determined  by the heat map responses. After integrating the 3D pose prior by the proposed approach, better pose estimates are obtained. \refFig{fig:mpii-bad} provides several failure examples. Visual inspection of these results suggests that the failures are mostly due to heavy occlusion, ambiguities from left-right symmetry, overlapping people, and extremely rare 3D poses that are beyond the representational capacity of the learned pose dictionary.

\subsection{Running time}

The experiments were performed on a desktop machine with an Intel i7 3.4G CPU, 8GB RAM and a GeForce GTX Titan X 6GB GPU.
The running times for CNN-based heat map generation (with the hourglass model) and convex initialization were roughly 0.3s and 0.6s per frame, respectively; both steps can be easily parallelized. The EM algorithm usually converged in 20 iterations with a CPU time less than 100s for a sequence of 300 frames.

The running time of our approach depends on the dictionary size. There is a trade-off between accuracy and efficiency.  For instance, for dictionary sizes of 32, 64, 96, and 128
tested on the first ``Directions'' sequence of S9, the mean reconstruction error (in mm) was 48.1, 46.0, 45.6 and 44.4, respectively, and the computation time (in seconds) was 91, 197, 317 and 488, respectively.

\section{Summary}

In summary, a 3D human pose estimation framework from a monocular image or video has been presented that consists of a novel synthesis between a deep learning-based 2D part regressor, a sparsity-driven 3D reconstruction approach, and a 3D temporal smoothness prior.  This joint consideration combines the discriminative power of state-of-the-art 2D part detectors,
the expressiveness of 3D pose models, and regularization by way of aggregating information over time.
In practice, alternative part detectors, pose representations, and temporal models can be conveniently integrated in the proposed framework by replacing the original components.  
Experiments demonstrated that 3D geometric priors and temporal coherence can not only help 3D reconstruction but also improve 2D part localization.

\appendix

\section*{Proof of Equation (19)}
For simplicity, $\mathcal{L}(\theta;\bfW)$ is rewritten as
\begin{align}
\mathcal{L}(\theta;\bfW) & = \sum_{t=1}^{n}\left\| \bfW_t - \bfR_t\sum_{i=1}^{k} c_{it}\bfB_i - \bfT_t\bfone^T \right\|_F^2 \nonumber \\
& =  \left\| \bfW-\bfM(\theta) \right\|_F^2,
\end{align}
where $\bfW$ is the stack of all $\bfW_t$ and $\bfM(\theta)$ is the stack of all $\bfR_t\sum_{i=1}^{k} c_{it}\bfB_i-\bfT_t\bfone^T$. Note that the constant $\frac{\nu}{2}$ is ignored for brevity. 

Then, we have:
\begin{align} \small
& \int \mathcal{L}(\theta;\bfW) \pr(\bfW|\bfI,\theta') d\bfW \nonumber \\
& =   \int \left\|\bfW-\bfM(\theta)\right\|_F^2 ~ \pr(\bfW|\bfI,\theta') d\bfW \nonumber \\
& =   \int \left\{ \|\bfW\|_F^2 - 2\left<\bfW,\bfM(\theta)\right> + \|\bfM(\theta)\|_F^2 \right\} ~ \pr(\bfW|\bfI,\theta') d\bfW \nonumber \\
& =   ~ \left\{ \mbox{const} - \int 2\left<\bfW,\bfM(\theta)\right> \pr(\bfW|\bfI,\theta') d\bfW + \|\bfM(\theta)\|_F^2  \right\} ~  \nonumber \\
& =   ~ \left\{ \mbox{const} - 2\left< \int\bfW\pr(\bfW|\bfI,\theta')d\bfW~,~\bfM(\theta)\right>  + \|\bfM(\theta)\|_F^2  \right\} ~  \nonumber \\
& =   \left\|\int\bfW\pr(\bfW|\bfI,\theta')d\bfW-\bfM(\theta)\right\|_F^2 + \mbox{const} \nonumber \\
& =   \left\|\mathrm{E}\left[\bfW|\bfI,\theta'\right]-\bfM(\theta)\right\|_F^2 + \mbox{const}
\end{align}
\vfill

\section*{Derivation of Equation (20)}
\begin{align}
\mathrm{E}\left[\bfW|\bfI,\theta'\right] & = \int \pr(\bfW|\bfI,\theta')~ \bfW ~ d\bfW \nonumber \\
& = \int \frac{\pr(\bfW,\bfI|\theta')}{\pr(\bfI|\theta')} ~ \bfW ~ d\bfW \nonumber \\
& = \int \frac{\pr(\bfI|\bfW)\pr(\bfW|\theta')}{M} ~ \bfW ~ d\bfW,
\end{align}
where $M$ is a constant.

\bibliographystyle{IEEEtran}
\bibliography{bibref_definitions_short,bibref}

\begin{IEEEbiography}[{\includegraphics[height=1.2in]{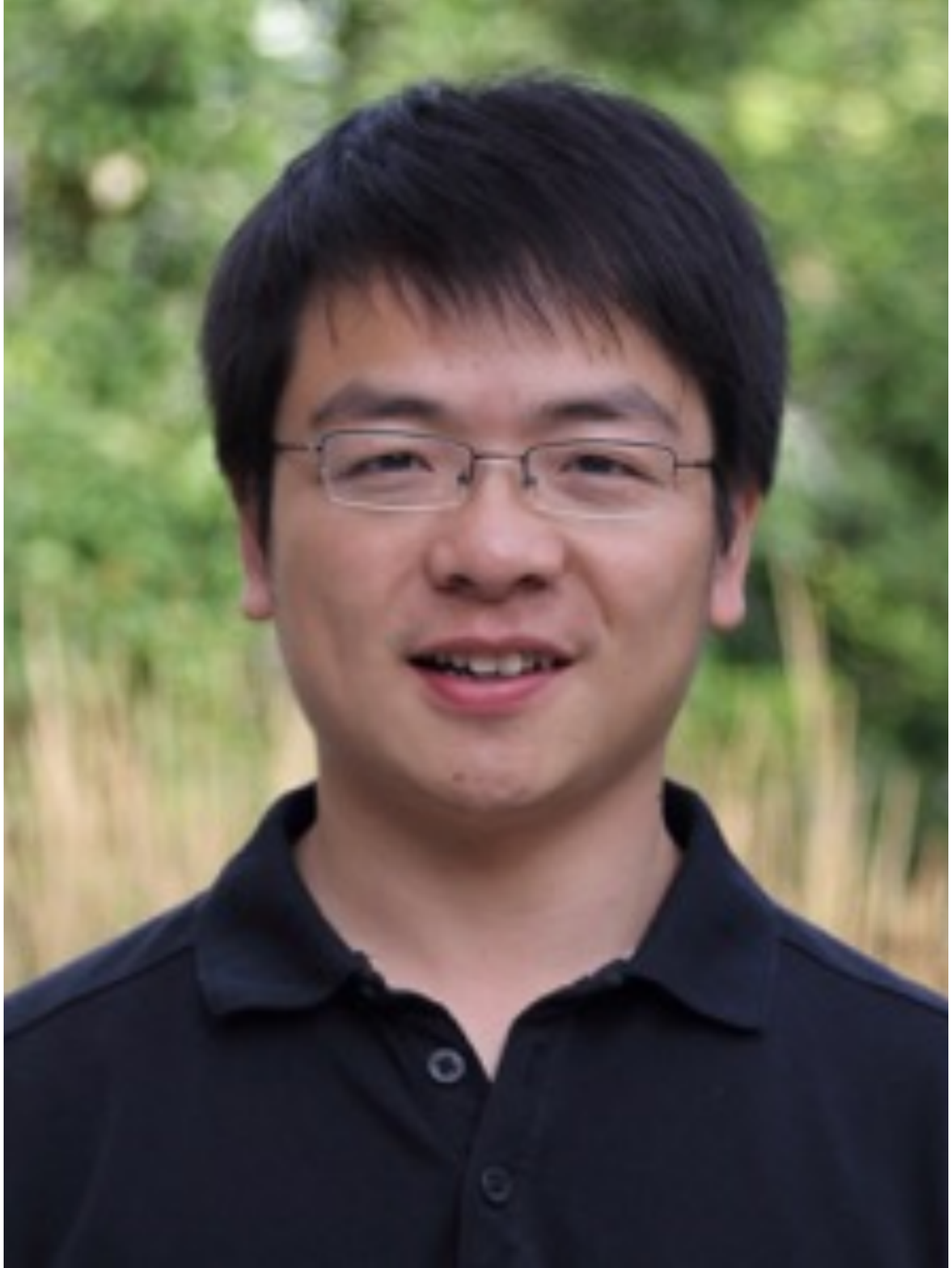}}]{Xiaowei Zhou}
is a Research Professor in the College of Computer Science, Zhejiang University. He was a Postdoctoral Researcher in Computer and Information Science, University of Pennsylvania. He obtained his Bachelor's degree in Optical Engineering from Zhejinag University, 2008, and his PhD degree in Electronic and Computer Engineering from The Hong Kong University of Science and Technology, 2013. His research interests are on 3D object recognition, human pose estimation, image matching, motion analysis and sparse/low-rank modeling.
\end{IEEEbiography}
\begin{IEEEbiography}[{\includegraphics[height=1.2in]{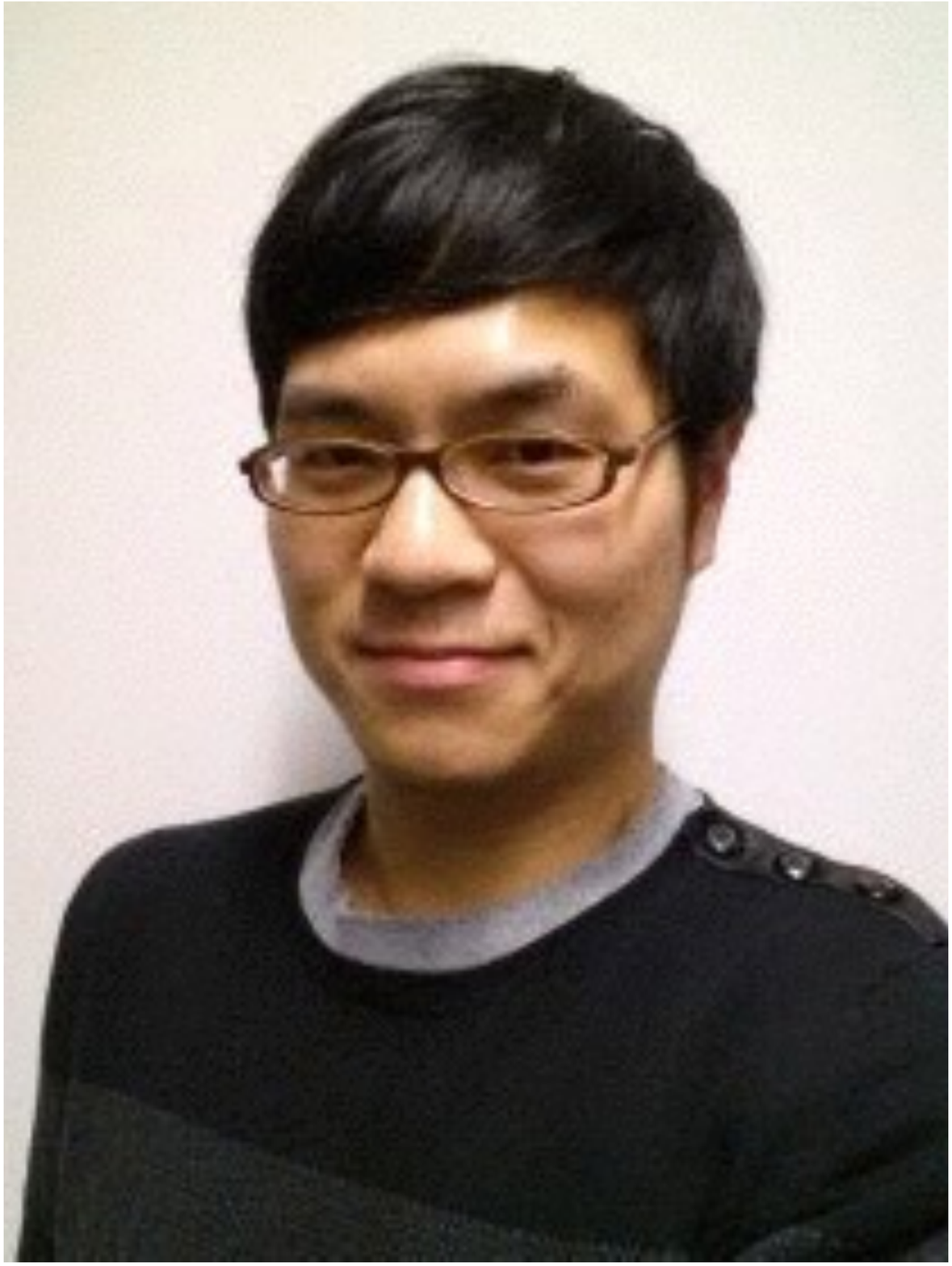}}]{Menglong Zhu}
is a Computer Vision Software Engineer at Google. He obtained a Bachelor's degree in Computer Science from Fudan University, in 2010, and a Master's degree in Robotics and a PhD degree in Computer and Information Science from University of Pennsylvania, in 2012 and 2016, respectively. His research interests are on object recognition, 3D object/human pose estimation, human action recognition, visual SLAM and text recognition.
\end{IEEEbiography}
\begin{IEEEbiography}[{\includegraphics[height=1.2in]{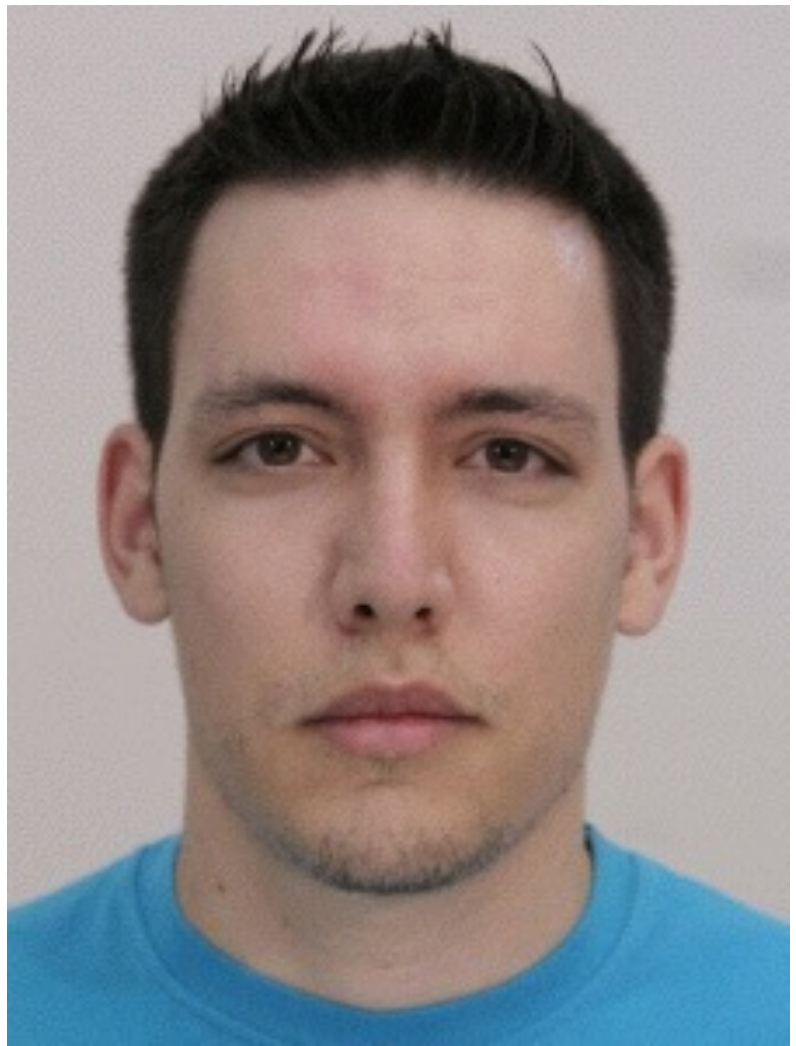}}]{Georgios Pavlakos}
is currently a doctoral student in Computer and Information Science, University of Pennsylvania. He received the BS degree in Electrical and Computer Engineering from the National Technical University of Athens, in 2014. His research interests lie at the intersection of computer vision and machine learning and include reconstruction and pose estimation of objects and humans from single images.
\end{IEEEbiography}
\begin{IEEEbiography}[{\includegraphics[height=1.2in]{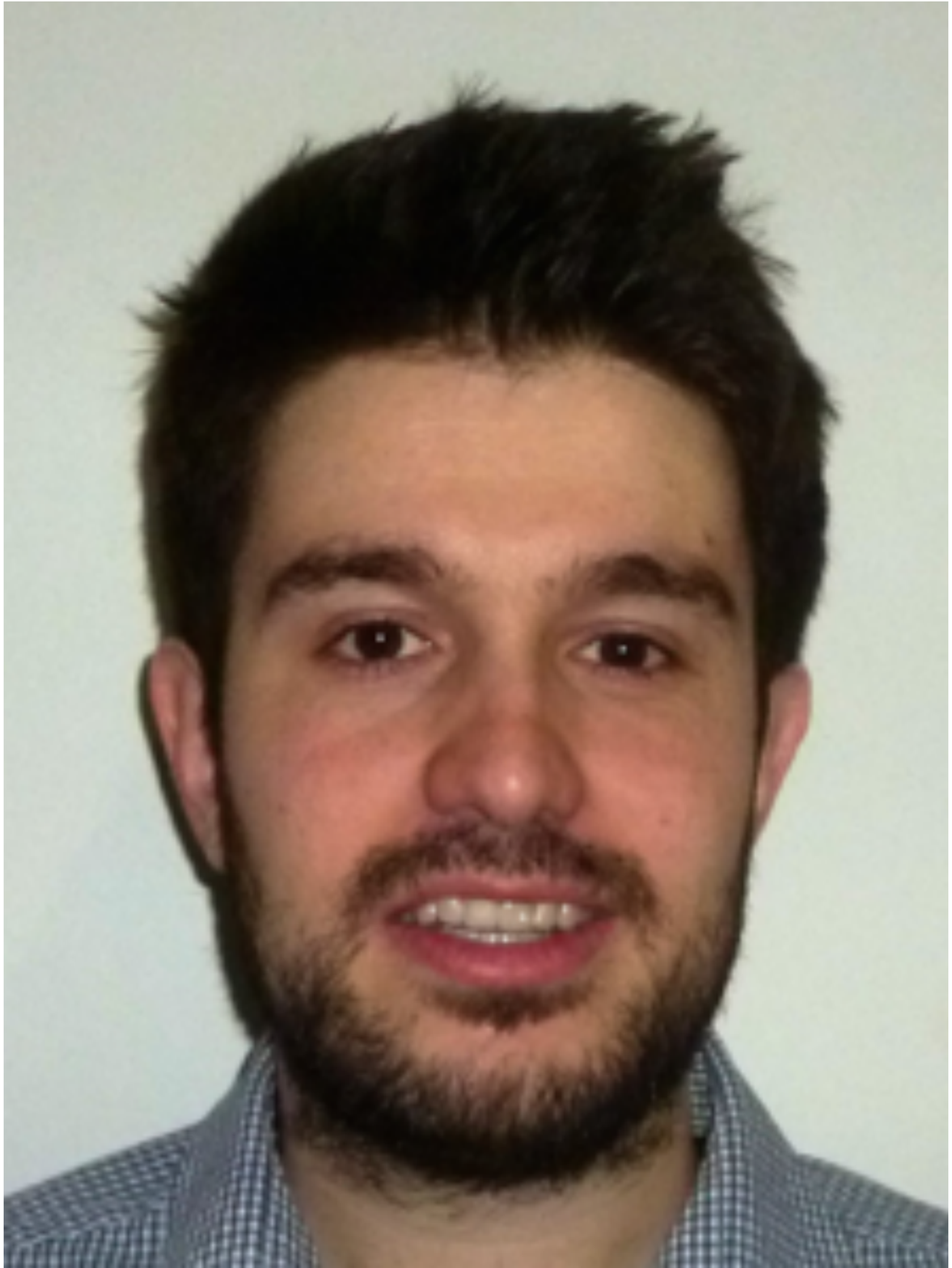}}]{Spyridon Leonardos}
is currently a doctoral student in Computer and Information Science, University of Pennsylvania. He  received the BS degree in Electrical and Computer Engineering (highest honors) from the National Technical University of Athens, in 2012 and the MS degree in Computer Science from University of Pennsylvania, in 2015. His research interests include  multiple view geometry,
reconstruction of articulated objects from video, Riemannian geometry for computer vision and sensor networks.
\end{IEEEbiography}
\begin{IEEEbiography}[{\includegraphics[height=1.2in]{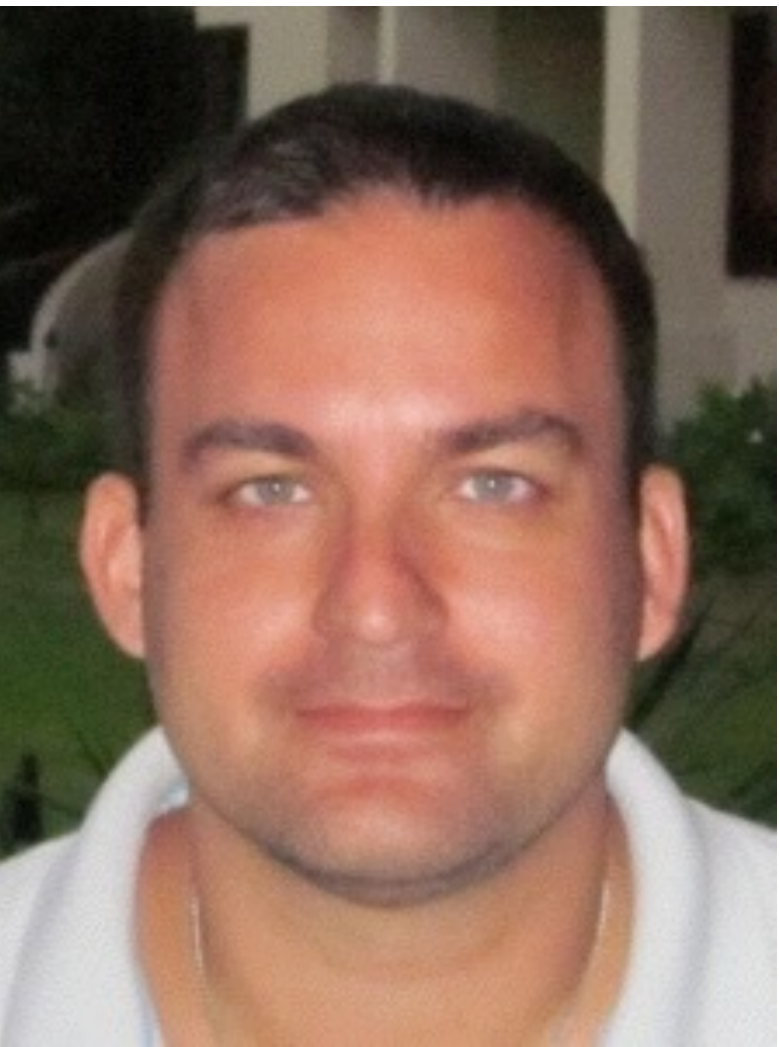}}]{Konstantinos G. Derpanis}
is an Associate Professor of Computer Science, Ryerson University, Toronto. He received the Honours Bachelor of Science (BSc) degree in Computer Science from the University of Toronto, in 2000, and the MSc and PhD degrees in Computer Science from York University, Canada, in 2003 and 2010, respectively. Subsequently, he was a postdoctoral researcher in the GRASP Laboratory at the University of Pennsylvania. His main research field of interest is computer vision with emphasis on motion analysis and human motion understanding, and related aspects in image processing and machine learning.
\end{IEEEbiography}
\begin{IEEEbiography}[{\includegraphics[height=1.2in]{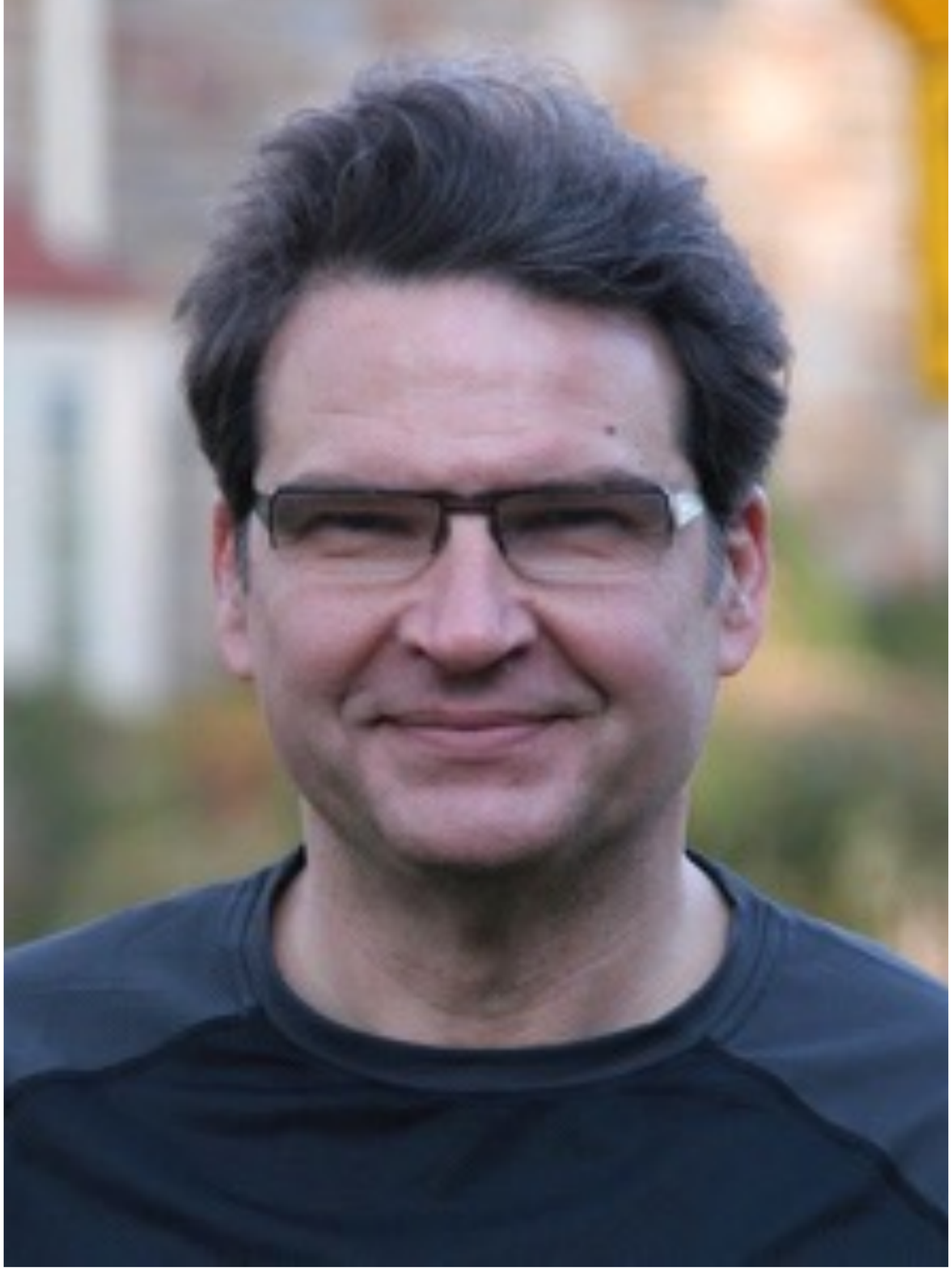}}]{Kostas Daniilidis}
is a Professor of Computer and Information Science, University of Pennsylvania. He obtained his MSE (Diploma) in Electrical
Engineering from the National Technical University of Athens, 1986, and his PhD (Dr.rer.nat.) in Computer Science from the University of Karlsruhe, 1992. He is an IEEE Fellow and served as Associate Editor of IEEE Transactions on Pattern Analysis and Machine Intelligence from 2003 to 2007. His research interests are on visual motion and navigation, active perception, 3D object detection and localization, and panoramic vision.
\end{IEEEbiography}
\vfill

\end{document}